\documentclass[10pt,reqno]{amsart}

\usepackage{hyperref}
\usepackage{amssymb}

\usepackage{amsmath}
\usepackage{amscd}
\usepackage{amsthm}

\usepackage{url}
\usepackage{mdwlist}
\usepackage[shortlabels]{enumitem}

\usepackage{xcolor}

\usepackage{float}

\usepackage{graphics}
\usepackage{epsfig}
\graphicspath{{./}{img/}}
\usepackage{subfig}

\usepackage[lined,boxed,linesnumbered,commentsnumbered,norelsize]{algorithm2e}

\usepackage{pgf}
\usepackage{tikz}
\usetikzlibrary{arrows,patterns,plotmarks,shapes,snakes,er,3d,automata,backgrounds,topaths,trees,petri,mindmap}

\usepackage{verbatim}			

\usepackage[T1]{fontenc}			

\usepackage{todonotes}			
\usepackage{marginnote}

\usetikzlibrary{calc}

\newif\ifnever\neverfalse

\newcommand{\bm}[1]{\mbox{\boldmath${#1}$}}

\newcommand{\domain}{\Omega}

\newcommand{\x}{\bm{x}}
\newcommand{\p}{\bm{p}}

\newcommand{\bs}{\bm{s}}
\newcommand{\bt}{\bm{t}}

\DeclareMathOperator*{\argmin}{argmin}

\newcommand{\bq}{\begin{equation}}
\newcommand{\eq}{\end{equation}}

\newcommand{\R}{\mathbb{R}} 

\newcommand{\Neigh}{\mathcal{N}}   
\newcommand{\Bd}{\mathcal{B}}      
\newcommand{\TT}{\mathcal{T}}      
\newcommand{\PF}{\mathcal{PF}}     
\newcommand{\Path}{P}              
\newcommand{\PathSet}{\mathcal{P}} 
\newcommand{\Graph}{\mathcal{G}}   

\newcommand{\abs}[1]{\left|{#1}\right|}              
\newcommand{\braces}[1]{\left[{#1}\right]}           
\newcommand{\cbraces}[1]{\left\{{#1}\right\}}        
\newcommand{\norm}[1]{\left\|{#1}\right\|}           
\newcommand{\parens}[1]{\left({#1}\right)}           
\newcommand{\ceil}[1]{\left\lceil{#1}\right\rceil}   
\newcommand{\myhat}[1]{\widehat{#1}}                 

\newcommand*{\CopyCounter}[2]{%
  \expandafter\def\csname c@#2\endcsname{\csname c@#1\endcsname}%
  \expandafter\def\csname p@#2\endcsname{\csname p@#1\endcsname}%
  \expandafter\def\csname the#2\endcsname{\csname the#1\endcsname}}

\numberwithin{Theorem}{section}
\CopyCounter{Theorem}{Proposition}
\CopyCounter{Theorem}{ProposedProblem}
\CopyCounter{Theorem}{Property}
\CopyCounter{Theorem}{Claim}
\CopyCounter{Theorem}{Lemma}
\CopyCounter{Theorem}{Corollary}
\CopyCounter{Theorem}{Conjecture}
\CopyCounter{Theorem}{Definition}
\CopyCounter{Theorem}{Example}
\CopyCounter{Theorem}{Remark}
\CopyCounter{Theorem}{Question}
\CopyCounter{Theorem}{Condition}
\CopyCounter{Theorem}{Criterion}
\CopyCounter{Theorem}{Observation}
\theoremstyle{plain}

\theoremstyle{definition}

\newif\ifnever\neverfalse

\newcommand{\marginfix}{
\setlength{\parskip}{0.01cm}
\setlength{\textwidth}{6.0in}
\setlength{\oddsidemargin}{-0.0 in}
\setlength{\evensidemargin}{0.0 in}
\setlength{\topmargin}{-0.5in}
\setlength{\textheight}{9.0 in}
}

\marginfix

  \renewenvironment{thebibliography}[1]{%
    \begin{oldthebibliography}{#1}%
      \setlength{\parskip}{.3ex}%
      \setlength{\itemsep}{.3ex}%
  }%
  {%
    \end{oldthebibliography}%
  }

\usepackage{tkz-euclide}

\begin{document}

{\Large{\bf
\centerline{A bi-criteria path planning algorithm for robotics applications.}
}}

\renewcommand*{\thefootnote}{\fnsymbol{footnote}} 
\vspace*{.15in}
{\Large
\centerline{
  Zachary Clawson\footnotemark[2], 
  Xuchu (Dennis) Ding\footnotemark[3], 
  Brendan Englot\footnotemark[4],
}
\vspace*{0.05in}
\centerline{
  Thomas A. Frewen\footnotemark[3], 
  William M. Sisson\footnotemark[3], 
  Alexander Vladimirsky\footnotemark[2]
}
}

\footnotetext[2]{Center for Applied Mathematics and Department of Mathematics, Cornell University, Ithaca, NY 14853. Both authors' work was supported in part by the National Science Foundation Grant DMS-1016150. (Emails: \texttt{zc227@cornell.edu} and \texttt{vlad@math.cornell.edu}.)}
\footnotetext[3]{United Technologies Research Center, Hartford, CT 06118. These authors' work was supported by United Technologies Research Center under the Autonomy Initiative. (Emails: \texttt{xuchu.ding@gmail.com}, \texttt{frewenta@utrc.utc.com}, and \texttt{sissonw2@utrc.utc.com}.)}
\footnotetext[4]{Schaefer School of Engineering \& Science, Stevens Institute of Technology, Hoboken, NJ 07030. (Email: \texttt{benglot@stevens.edu}.)}

\renewcommand*{\thefootnote}{\arabic{footnote}} 

\vspace*{.15in}
\noindent
{\small
{\bf Abstract: }
Realistic path planning applications often require optimizing with respect to several criteria simultaneously.
Here we introduce an efficient algorithm for bi-criteria path planning on graphs.
Our approach is based on augmenting the state space to keep track of the ``budget'' remaining to satisfy the constraints on secondary cost.
The resulting augmented graph is acyclic and the primary cost can be then minimized by a simple upward sweep through budget levels.
The efficiency and accuracy of our algorithm is tested on Probabilistic Roadmap graphs to minimize the distance of travel subject to a constraint on the overall threat exposure of the robot.  We also present the results from field experiments illustrating the use of this approach on realistic robotic systems.
}

\vspace*{.15in}

\section{Introduction}
\label{s:intro}

The shortest path problem on graphs is among the most studied problems of computer science, with numerous applications ranging from robotics to image registration.
Given the node-transition costs, the goal is to find the path minimizing the cumulative cost from a starting position up to a goal state. 
When there is a single (nonnegative) cost this can be accomplished efficiently by Dijkstra's algorithm \cite{dijkstra1959note} and a variety of related {\em label-setting methods}.

However, in many realistic applications paths must be evaluated and optimized according to several criteria simultaneously (time, cumulative risk, fuel efficiency, etc).  Multiple criteria lead to a natural generalization of path optimality: a path is said to be
(weakly) {\em Pareto-optimal} if it cannot be improved according to all criteria simultaneously.
For evaluation purposes, each Pareto-optimal path can be represented by a single point, whose coordinates are cumulative costs with respect to each criterion. 
A collection of such points is called the {\em Pareto Front} (PF), which is often used by decision makers in a posteriori evaluation of optimal trade-offs.  
A related practical problem (requiring only a portion of PF) is to optimize with respect to one (``primary'') criterion only, but with upper bounds enforced based on the remaining (``secondary'') criteria.

It is natural to approach this problem by reducing it to single criterion optimization: a new transition-cost is defined as a weighted average of transition penalties specified by all criteria, and Dijkstra's method is then used to recover the ``shortest'' path based on this new averaged criterion.  
It is easy to show that the resulting path is always Pareto-optimal for every choice of weights used in the averaging. Unfortunately, this {\em scalarization approach} has a significant drawback \cite{Das}: it finds the paths corresponding to convex parts of PF only, while the non-convexity of PF is quite important in many robotics applications. 
The positive and negative algorithmic features of scalarization are further discussed in \S \ref{ss:scalarization_comparison}.

The same problem applied to a stochastic setting in the context of 
a Markov Decision Processes (MDP) is called stochastic shortest path 
problem, which can be seen as a a generalization of the deterministic 
version considered in this paper.  In this case multiple costs can be 
considered such that a primary cost is optimized, and an arbitrary 
number of additional costs can be used as constraints.  Such problem 
can generally be solved as a Constrained MDP (CMDP) \cite{altman1999constrained}, and recent work 
to extend this approach is ``Chance-constrained'' CMDP \cite{OnoPavoneCcDp} that works well for large state spaces \cite{FeyzHsCmdp,DingCmdp} for robotics applications.  However, by 
considering the additional costs as constraints, the PF cannot be 
extracted.

Numerous generalizations of label-setting methods have also been developed to recover {\em all} Pareto optimal paths
\cite{Jaffe, Stewart, TungChew, MandowPerez1, MandowPerez2, SkriverAndersen, Machuca_evaluation, Machuca_comparison}.  
However, all of these algorithms were designed for general graphs, and their efficiency advantages are less obvious for highly-refined geometrically embedded graphs, where Pareto-optimal paths are particularly plentiful.  
(This is precisely the case for meshes or random-sampling based graphs intended to approximate the motion planning in continuous domains.)  In this paper we describe a simple method for bi-criteria path planning on such graphs, with an efficient approach for approximating the entire PF.  
The key idea is based on keeping track of the ``budget'' remaining to satisfy the constraints based on the secondary criterion.  This approach was previously used in continuous optimal control by Kumar and Vladimirsky \cite{KumarVlad} to approximate the discontinuous value function on the augmented state space.  
More recently, this technique was extended to hybrid systems modeling constraints on reset-renewable resources \cite{Unsafe_Sets}.  
In the discrete setting, the same approach was employed as one of the modules in hierarchical multi-objective planners \cite{ding2014hierarchical}; our current paper extends that earlier conference publication.

The key components of our method are discussed in \S \ref{s:algo}, followed by the implementation notes in \S \ref{s:implement}.
In \S \ref{s:bench} we provide the results of numerical tests on Probabilistic RoadMap graphs (PRM) in two-dimensional domains with complex geometry.  Our algorithms were also implemented on a realistic heterogeneous robotic system and tested in field experiments described in \S \ref{s:experiment}.  Finally, open problems and directions for future work are covered in \S \ref{s:conclusions}.

\section{The augmented state space approach}
\label{s:algo}

\ifnever
{\em This is still a placeholder -- essentially the first part of
the ``pseudocodes document'' that I previously wrote for UTRC.
It might be better to present everything on general graphs, postponing the discussion of PRM.
The secondary costs are starting out quantized in multiples of $\delta$; the same $\Delta b = \delta$ is then also used as the
distance between adjusted budget levels.  To be changed since in practice we almost always use $\Delta b > \delta$.
Also, here everything is expanded from the target whereas the implemented version expands from the source.}
\fi

Consider a directed graph $\Graph$ on the set of nodes $X=\{\x_1, \ldots, \x_n, \x_{n+1} = \bs \}$ and edges $E$ between nodes. 
We will choose $\bs$ to be a special ``source'' node (the robot's current position) and our goal will be to find optimal paths starting from $\bs$.  For convenience, we will also use the notation $\Neigh(\x_j) = \Neigh_j$ for the set of all nodes $\x_i$ from which there exists a direct transition to $\x_j$.  
We will assume that the cost of each such transition $C_{ij}$ is positive. 
We begin with a quick review of the standard methods for the classical shortest path problems.

\subsection{The single criterion case.}
Using $\Phi(P)$ to denote the cumulative cost (i.e., the sum of transition costs $C_{ij}$'s) along a path $P$,
the usual goal is to minimize $\Phi$ over $\PathSet_j$, the set of all paths from $\bs$ to $\x_j \in X$.
The standard dynamic programming approach is to introduce the {\em value function} $U(\x_j)=U_j$, describing the total cost along any such optimal path.  Bellman's optimality principle yields the usual coupled system of equations:
\begin{equation}\label{Bellman_primary}
U_j \; = \; \min_{\x_i \in \Neigh_j} \cbraces{ C_{ij} \ + \ U_i  }, \qquad \forall \ j =1,2,\ldots, n
\end{equation}
with $U_{n+1} = U(\bs) = 0$.
Throughout the paper we will take the minimum over an empty set to be $+\infty$.
The vector $U = (U_1, \ldots, U_n)$ can be formally viewed as a fixed point of an operator
$\TT: \mathbb{R}^n \to \mathbb{R}^n$ defined componentwise by \eqref{Bellman_primary}.
A naive approach to solving this system is to use {\em value iteration}:  starting with an overestimate initial guess
$u_0 \in \mathbb{R}^n$, we could repeatedly apply $\TT$, 
obtaining the correct solution vector $U$ in at most $n$ iterations.  This results in $O(\kappa n^2)$ computational cost
as long as the in-degrees of all nodes are bounded: $\abs{\Neigh_j} < \kappa$ for some constant $\kappa \ll n$.
The process can be further accelerated using the Gauss-Seidel relaxation, but then the number of iterations will
strongly depend on the ordering of the nodes within each iteration.  For acyclic graphs, a standard topological ordering of nodes can be used to essentially decouple the system of equations. In this case, the Gauss-Seidel relaxation converges after a single iteration, yielding the $O(n)$ computational complexity.  For a general directed graph, such a {\em causal} ordering of nodes is a priori unknown, but can be recovered at runtime by exploiting the monotonicity of $U$ values along every optimal path.  This is the essential idea behind Dijkstra's classical algorithm \cite{dijkstra1959note}, which relies on heap-sort data structures and solves this system in $O(n \log n)$ operations.  Alternatively, a class of {\em label-correcting} algorithms (e.g., \cite{bertsekas1993SLF, bertsekas1996LLL, bertsekas1995dynamic}) attempt to approximate the same causal ordering, while avoiding the use of heap-sort data structures.  These algorithms still have the $O(n^2)$ worst-case complexity, but in practice are known to be at least as efficient as Dijkstra's on many types of graphs \cite{bertsekas1993SLF}.

When we are interested in optimal directions from $\bs$ to a specific node $\bt \in X$ only,  Dijkstra's method can be terminated as soon as $U(\bt)$ is computed.  (Note: Dijkstra's algorithm typically expands computations outward from $\bs$.)  An A* method \cite{hart1968formal} can be viewed as a speed-up technique, which further restricts the computational domain (to a neighborhood of $(\bs,\bt)$ optimal path) using heuristic underestimates of the cost-to-go function.
More recent extensions include ``Anytime A*'' (to ensure early availability of good suboptimal paths)
\cite{hansen1997anytime, likhachev2003ara, van2011ana} and
a number of algorithms for dynamic environments (where some of the $C_{ij}$ values might change before we reach $\bt$) \cite{stentz1994optimal, stentz1995focused, koenig2005fast}.

\subsection{\bf Bi-criteria optimal path planning: different value functions and their DP equations.} 
We will now assume that an alternative criterion for evaluating path quality is defined by specifying ``secondary costs'' $c_{ij}>0$ for all the transitions in this graph.  Similarly, we define $\phi(\Path)$ to be the cumulative secondary cost (based on $c_{ij}$'s) along $\Path$.
In this setting, it is useful to consider several different value functions.
The secondary (unconstrained) value function is $V_j = \min\limits_{\Path \in \PathSet_j}\phi(\Path).$   It satisfies a similar system of equations:
$V_{\bs} = 0$ and
\[
V_j = \min_{\x_i \in \Neigh_j} \cbraces{ c_{ij} + V_i }, \quad j \leq n.
\]

A natural question to ask is how much cost must be incurred to traverse a path selected to optimize a {\em different} criterion.
We define $\tilde{V}_j$ as $\phi(\Path)$ minimized over the set of all primary-optimal paths $\cbraces{\Path \in \PathSet_j \mid \Phi(\Path) = U_j}.$
Thus, $\tilde{V}_{\bs} = 0$, and if we define $\Neigh_j' = \argmin\limits_{\x_i \in \Neigh_j} \cbraces{C_{ij} + U_i} \subset \Neigh_j$, then
\[
\tilde{V}_j = \min\limits_{\x_i \in \Neigh_j'} \cbraces{ c_{ij} + \tilde{V}_i }, \quad j \leq n.
\]
Similarly, we define $\tilde{U}_j$ as $\Phi(P)$ minimized over the set of all secondary-optimal paths $\cbraces{\Path \in \PathSet_j \mid \phi(\Path) = V_j}.$ 
Thus, $\tilde{U}_{\bs} = 0$, and if we define $\Neigh_j'' = \argmin\limits_{\x_i \in \Neigh_j} \cbraces{c_{ij} + V_i}
 \subset \Neigh_j$, then
\[
\tilde{U}_j = \min\limits_{\x_i \in \Neigh''_j} \cbraces{C_{ij} + \tilde{U}_i }, \quad j \leq n.
\]
All $U_j$'s can be efficiently obtained by the standard Dijkstra's method
with $\tilde{V}_j$'s computed in the process as well.
The same is true for $V_j$'s and $\tilde{U}_j$'s.

Returning to our main goal, we now define the primary-constrained-by-secondary value function $W(\x_j, b) = W_j^b$
as $\Phi(\Path)$ (the accumulated primary cost) minimized over $\cbraces{P \in \PathSet_j \mid \phi(\Path) \leq b }$
(the paths with the cumulative secondary cost not exceeding $b$).  
We will say that $b$ is {\em the remaining budget} to satisfy the secondary cost constraint.
This definition and the positivity of secondary costs yield several useful properties of $W$:
\begin{enumerate}
\item $W_j^b$ is a monotone non-increasing function of $b$.
\item $b < V_j \; \Longleftrightarrow \; W_j^b = +\infty.$
\item $b = V_j \; \Longleftrightarrow \; W_j^b = \tilde{U}_j.$
\item $b \geq \tilde{V}_j \; \Longleftrightarrow \; W_j^b = U_j.$
\end{enumerate}

We will use the notation
\bq
\label{beta_simple}
\beta(i,b,j) \; = \; b \, - \, c_{ij}
\eq
to define the set of feasible transitions on level $b$:
\[
\Neigh_j(b) = \{ \x_i \in \Neigh_j \mid \beta(i,b,j) \geq 0 \}.
\]
The dynamic programming equations for $W$ are then
$W_{\bs}^b = 0$ for all $b$ and
\begin{equation}
\label{DP:extended}
W_j^b = \min\limits_{\x_i \in \Neigh_j(b)}
\cbraces{ C_{ij} + W_i^{\beta(i,b,j)}}, \qquad j \leq n, \, b>0.
\end{equation}
For notational simplicity, it will be sometimes useful
to have $W$ defined for non-positive budgets;
in such cases we will assume $W_j^b = +\infty$ whenever $b \leq 0$ and $j \leq n$. 
The Pareto Front for each node $\x_j$ is the set of pairs $\parens{b,W_j^b}$, where a new reduction in the primary cost is achieved:
\begin{equation}\label{eq:PF_definition}
\PF_j = \cbraces{\parens{b,W_j^b} \, \mid \,  W_j^b < W_j^{b'}, \,\forall \, b' < b}.
\end{equation}
For each fixed $\x_j$, the value function $W_j^b$ is
piecewise-constant on $b$-intervals and the entries
in $\PF_j$ provide the boundaries/values for these intervals.

Given the above properties of $W$, it is only necessary to solve the system \eqref{DP:extended} for $b \in [0, B]$,
where $B$ is the  {\em maximal budget level}.  
E.g., for many applications it might be known a priori 
that a secondary cumulative cost above some $B$ is unacceptable.  
On the other hand, if the goal is to recover the entire $\PF_j$ for a specific $\x_j \in X$, we can choose 
$B = \tilde{V}_j$, or even $B = \max_i \tilde{V}_i$ to accomplish this for all $\x_j$.

Since all $c_{ij}$ are positive, we observe that the system \eqref{DP:extended} is {\em explicitly causal}:
the expanded graph on the nodes $\cbraces{\x_j^b \mid \x_j \in X, b \geq 0 }$ is acyclic and the causal ordering of
the nodes is available a priori.  The system \eqref{DP:extended} can be solved by a single Gauss-Seidel sweep
in the direction of increasing $b$.

\subsection{The basic upward-sweep algorithm.} 

For simplicity, we will first assume that all secondary costs are positive and quantized:
\[
\exists \, \delta > 0 \; \text{ s.t.  all } c_{ij} \, \in \, \cbraces{\delta, 2\delta, 3\delta, \ldots}.
\]
This also implies that $W_j^b$ is only defined for $b \in \Bd = \cbraces{0, \delta, 2\delta, \ldots, B := m \delta }.$
A simple implementation (Algorithm \ref{alg:DSP_onesweep}, described below) takes advantage of this structure by storing $W_j^b$ as an array of values for each node $\x_j \in X \setminus \cbraces{\bs}$ and $b \in \Bd$. 
For $\bs$ we will need only one value $W_{\bs} = W_{\bs}^b = 0$ for all $b \in \Bd$. For each remaining node $\x_j \in X \setminus \{\bs\}$ with budget $b \in \Bd$ we can compute the primary-constrained-by-secondary value function $W_j^b$ using formula \eqref{DP:extended}.

\begin{algorithm}
\SetKwInOut{Compute}{compute}
\SetKwInOut{Init}{Initialization}
\Init{}
\BlankLine
Compute $U, V, \tilde{U}, \tilde{V}$ for all nodes by Dijkstra's method. \\
\BlankLine
Set $W_{\bs} := 0$\\

\vspace{0.4cm}

\textbf{Main Loop:}\\
\BlankLine
	\ForEach{$b = \delta, \ldots, B$} {

		\ForEach{$\x_j \in X \backslash \{\bs\}$} {
			
			\BlankLine
			\eIf{($U_j < \infty$) AND ($b \geq V_j$)} {
				\BlankLine
				\eIf{$b = V_j$} {
					\BlankLine
					$W_j^b := \tilde{U}_j$;
					\BlankLine
				}{
					\eIf{$b < \tilde{V}_j$} {
						\BlankLine
						Compute $W_j^b$ from equation \eqref{DP:extended};\\
						\BlankLine
					}{
						\BlankLine
						$W_j^b := U_j$;
						\BlankLine
					}
				}
			}{
				$W_j^b := + \infty.$
			}
			\BlankLine

		}
	}

\BlankLine
\BlankLine
\caption{The basic explicitly causal (single-sweep) algorithm.
}
\label{alg:DSP_onesweep}
\end{algorithm}

The explicit causality present in the system is taken advantage of by the algorithm, leading to at most $\abs{X} \cdot \abs{ \Bd } = n m$ function calls to solve equation \eqref{DP:extended}. This results in an appealing $O(\kappa n m)$ complexity, linear in the number of nodes $n$ and the number of discrete budget levels $m$.

\subsection{Quantizing secondary costs and approximating  $\PF$}

In the general case, the secondary costs need not be quantized, and even if they are, the resulting number of budget levels $m$ might be prohibitively high for online path-planning.
But a slight generalization of Algorithm \ref{alg:DSP_onesweep} is still applicable to produce a conservative approximation of $W$ and $\PF$.  
This approach relies on ``quantization by overestimation'' of secondary edge weights with a chosen $\delta>0$:
\begin{equation}\label{eq:quantization_of_c}
\myhat{c}_{ij} \; := \; \delta \ceil{ \frac{c_{ij}}{\delta} } \; \geq \; c_{ij}.
\end{equation}
Similarly, we can define $\hat{\phi}(P)$ to be the cumulative quantized secondary cost along $P$. 

The definition of $\beta$ in \eqref{beta_simple} is then naturally modified to
\begin{equation}
\label{beta_conservative}
\beta(i,b,j) \; = \; b \, - \, \myhat{c}_{ij},
\end{equation}
with the set of feasible transitions on level $b \in \Bd$ still defined as
$\Neigh_j(b) = \{ \x_i \in \Neigh_j \mid \beta(i,b,j) \geq 0 \}.$
Modulo these changes, the new value function $\myhat{W}_j^b$ is also defined by \eqref{DP:extended} for all $b \in \Bd$.
It can be similarly computed by Algorithm \ref{alg:DSP_onesweep}  if  
the condition on line 7 is replaced by\\
\centerline{\tt if $b \in [V_j, \, V_j+ \delta)$ then ...}

The above modifications ensure that $\hat{\phi} \geq \phi$ for every path and $W_j^b \leq \myhat{W}_j^b$ for all $b \in \Bd$, $\x_j \in X.$
Moreover,  if $\myhat{W}_j^b$ is finite, there always exists some ``optimal path'' $P \in \PathSet_j$ such that 
$\hat{\phi}(P) \leq b$ and $\Phi(P)= \myhat{W}_j^b$.
Of course, there is no guarantee that such a path $P$ is truly Pareto-optimal with respect to the real (non-quantized) $c_{ij}$ values, but $\hat{\phi}(P)  \to \phi(P)$ and the obtained $\myhat{\PF}_j$ converges to the non-quantized Pareto Front as $\delta \to 0$. 
In fact, it is not hard to obtain the bound on  $\hat{\phi}(P)  - \phi(P)$ by using the upper bound on the number of transitions in Pareto-optimal paths. 
Let $c = \min_{i,j} c_{ij}$ and $C = \min_{i,j} C_{ij}$.
If $k(\Path)$ is the number of transitions on some Pareto-optimal path $\Path$ from $\bs$ to $\x_j$,
then 
\[
k(\Path) \; \leq \; K \, := \, \min \left(
\frac{ \tilde{V}_j}{c}, \,
\frac{ \tilde{U}_j}{C}
\right).
\]
Since Algorithm \ref{alg:DSP_onesweep} overestimates the secondary cost of each transition by at most $\delta$, we know that 
\[
\phi(P) \; \geq \; \hat{\phi}(P) \, - \, K \delta.
\]
In the following sections we show that much more accurate ``budget slackness'' estimates can be also produced at runtime.

\section{Implementation notes}
\label{s:implement}


A discrete multi-objective optimization problem is defined by the choice of domain $\Omega$, the graph $\Graph$ encoding allowable (collision-free) transitions in $\Omega$, and the primary and secondary transition cost functions $C$ and $c$ defined on the edges of that graph. 
Our application area is path planning for a ground robot traveling toward a goal waypoint $\x_j \in X$ in the presence of enemy threat(s). 
The Pareto Front consists of pairs of threat-exposure/distance values (each pair corresponding to a different path), where any decrease in threat exposure results in an increase in distance traveled and vice-versa.

While recovering the Pareto Front is an important task, deciding how to use it in a realistic path-planning application is equally important. 
If $c$ is based on the threat exposure and the maximal allowable exposure $B \geq V_j$ is specified in advance, 
the entire $\PF_{j}$ is not needed.  Instead, the algorithm can be used in a fully automatic regime, selecting the path corresponding to 
$W^B_{j}$.  
Alternatively, choosing $B = \tilde{V}_j$ we can recover the entire Pareto Front, and a human expert can then analyze it to select the best trade-off between $\Phi$ and $\phi$.  This is the setting used in our field experiments described in Section \ref{s:experiment}.

\subsection{Discretization parameter $\delta$.}
\label{ss:delta_implementation}

After the designation of $\Omega$, $C$, and $c$, the only remaining parameter needed for Algorithm 1 is $\delta > 0$. 
For a fixed graph $\Graph$, the choice of $\delta$ 
strongly influences the performance:
\setlist[enumerate,1]{leftmargin=0.7cm}
\begin{enumerate}[1.]
\item The selection of $\delta$ provides direct control over the runtime of the algorithm on $\mathcal{G}$.
Due to the fact that our robot's sampling-based planning process produces graphs of a predictable size, a fixed number of secondary budget levels $m$ ensures a predictable amount of computation time. 
Since the complexity is linear in $m$, the user can use the largest affordable $m$ to define
\begin{equation}\label{eq:delta_M}
\delta
\ := \
\tilde{V}_j \, / \, m,
\end{equation}
where $\tilde{V}_j$ is based on the \textsl{non-quantized} secondary weights $c$. 
\item 
The size of $\delta$ also controls the coarseness of the approximate Pareto Front.  After all, $\left| \myhat{\PF}_j \right| \leq m = B / \delta$, and
the approximation is guaranteed to be very  coarse if the number of Pareto-optimal paths in $\PathSet_j$ is significantly higher.
\item 
Even if $m$ is large enough to ensure that the number of Pareto optimal paths is captured correctly, there may be still an error in approximating 
$\PF_j$ due to quantization of secondary costs. 
For each pair $(b, \myhat{W}_j^b) \in \myhat{\PF}_j$ there is a corresponding path $\Path_j^b \in \PF_j$, whose
\textsl{budget slackness} can be defined as  
\begin{equation}\label{eq:slackness_definition}
S_j^b
\ = \ S\parens{\Path_j^b} 
\ = \ \hat{\phi}\parens{\Path_j^b} \, - \, \phi\parens{\Path_j^b} 
\ = \ b \, - \, \phi\parens{\Path_b}
\ \geq \ 0,
\end{equation}
which can be considered a post-processing technique. 
Alternatively, if the last transition in this path is from $\x_i$ to $\x_j$, this can be used to recursively define the slackness measurement
$$
S^b_j \; = \; \myhat{c}_{ij} \, - \, c_{ij} \, + \, S^{b \, - \, \myhat{c}_{ij}}_i,
$$ 
and compute it simultaneously with $\myhat{W}_j^b$ as an error estimate for $\myhat{\PF}_j$.  
This approach was introduced in \cite{Rhoads}.
\end{enumerate}
\noindent%
We point out two possible algorithmic improvements not used in our current implementation:
\begin{enumerate}[1.]
\item As described, the memory footprint of our algorithm is proportional to $n B / \delta$ since all $W$ values are stored on every budget level.
In principle, we need to store only finite values larger than $U_j$;  for each node $\x_j$, their number is $m(j)=(\tilde{V}_j - V_j)/\delta$.
This would entail obvious modifications of the main loop of Algorithm \ref{alg:DSP_onesweep}
since we would only need to update the ``still constrained'' nodes:
\[
\text{Still\_Constrained}(b) = \cbraces{\x_j \in X \backslash \{\bs\} \mid b \in \left[ V_j, \tilde{V}_j \right) }.
\]
This set can be stored as a linked list and efficiently updated as $b$ increases, especially if all nodes are pre-sorted based on $\tilde{V}_j$'s.
\item If the computational time available only permits for a coarse budget quantization (i.e. $\delta$ is large), an initial application of Algorithm \ref{alg:DSP_onesweep} may result in failure to recover a suitable number of paths from a set of evenly-spaced budget levels. 
Large differences in secondary cost often exist between paths that lie in different homotopy classes, and individual homotopy classes may contain a multitude of paths with small differences in secondary cost. 
To remedy this, $\delta$ might be refined adaptively, for a specific range of budget values.  The slackness measurements can be employed to determine when such a refinement is necessary.
\end{enumerate}

\subsection{Non-monotone convergence: a case study.}
\label{ss:nonmonotone_conv}

In real-world path planning examples, the true secondary transition costs $c_{ij}$ are typically not quantized, and equation \eqref{eq:quantization_of_c} introduces a small amount of error dependent on $\delta$. 
These errors will vanish as $\delta\to0$, but somewhat  counterintuitively the convergence is generally not monotone. 

\begin{figure}
\begin{center}
\subfloat[Original graph]{\label{fig:graph_quant:original}
\begin{tikzpicture}[point/.style={draw,shape=circle,inner sep = 1mm,minimum size = .25cm},scale = 0.75,->,>=stealth',shorten >=1pt,shorten <=1pt,auto,node distance = 1.5cm,semithick]
\begin{scope}[shift={(0,0)}, scale=0.9]
\draw[-] (-1.5,-1.75) rectangle (5.5,1.0);
\node[point,fill=white] (X0) at (0,0) {$\scriptstyle{x_0}$};
\node[point,fill=white] (X1) at (2,0) {$\scriptstyle{x_1}$};
\node[point,fill=white] (X2) at (4,0) {$\scriptstyle{x_2}$};
\draw (X0) edge[color=blue!60, line width=1.4pt] node [above] {$\scriptstyle{1.4}$} (X1);
\draw (X1) edge[color=blue!60, line width=1.4pt] node [above] {$\scriptstyle{1.4}$} (X2);
\draw (X0) edge[dashed,out=315,in=225,color=red!60] node [below] {$\scriptstyle{3}$} (X2);
\node[] at (0,-4.25) {};
\end{scope}
\end{tikzpicture}
} \hspace{4mm}
\subfloat[Cost on top path]{%
	\label{fig:graph_quant:blue}%
	\includegraphics[scale=0.35]{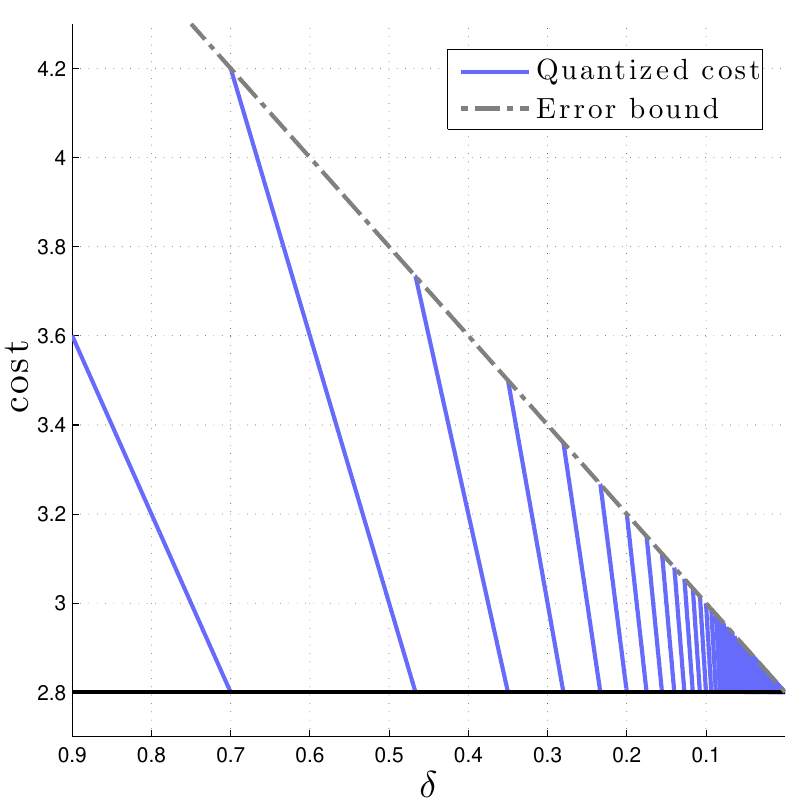}%
	}
\subfloat[Cost on bottom path]{%
	\label{fig:graph_quant:red}%
	\includegraphics[scale=0.35]{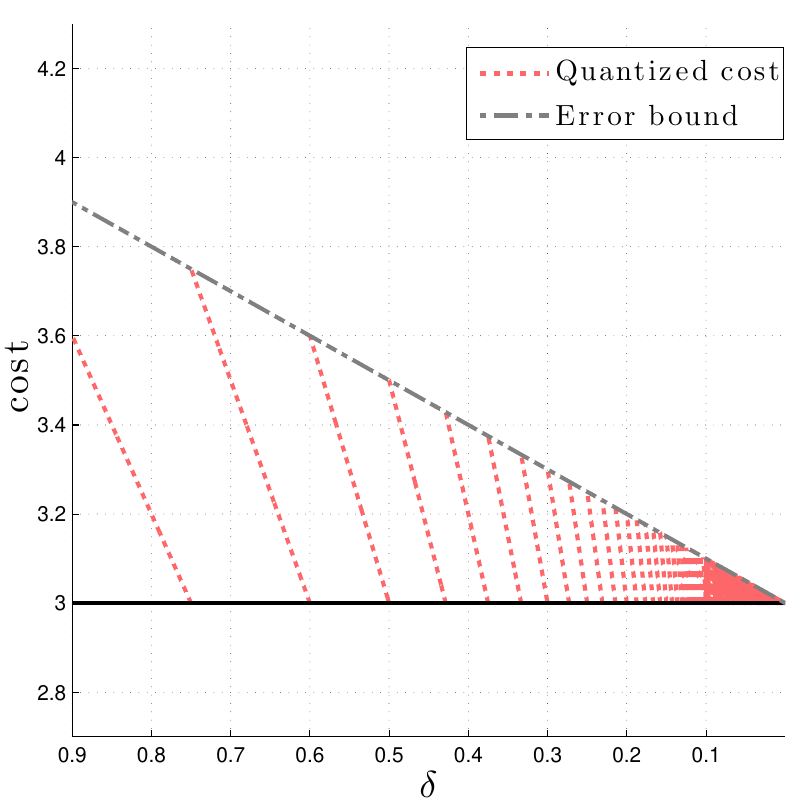}%
	}
\end{center}
\caption{
\small 
\textsc{(a)} Graph with two feasible paths from $\x_0$ to $\x_2$.
\textsc{(b) \& (c)} Quantized cost along the top and bottom paths (respectively) as $\delta$ decreases (left-to-right). The black lines represent the true cost in each figure (2.8 and 3, respectively).
}
\label{fig:graph_quant}
\end{figure}
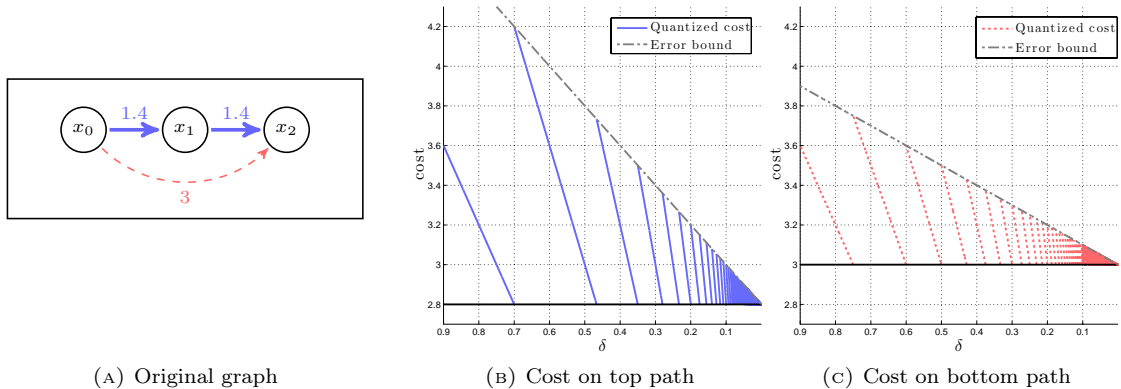

This issue can be illustrated even in single criterion path-planning problems.
Consider a simple graph in Fig. \ref{fig:graph_quant:original}, where there are two paths to travel from $\x_0$ to $\x_2$.  
If the edge-weights for this graph were quantized with formula \eqref{eq:quantization_of_c}, an upper bound for a given path would be:
\[
\hat{\phi}(\Path) \; \leq \; \phi(\Path) \, + \, \delta \cdot k(\Path).
\]

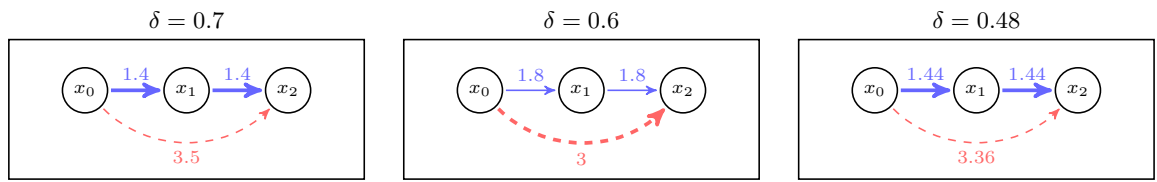
\begin{figure}
\begin{center}
\begin{tikzpicture}[point/.style={draw,shape=circle,inner sep = 1mm,minimum size = .25cm},scale = 0.75,->,>=stealth',shorten >=1pt,shorten <=1pt,auto,node distance = 1.5cm,semithick]

\begin{scope}[shift={(0,0)}, scale=0.9]
\draw node at (2,1.4) {\small$\delta=0.7$};
\draw[-] (-1.5,-1.75) rectangle (5.5,1.0);
\node[point,fill=white] (X0) at (0,0) {$\scriptstyle{x_0}$};
\node[point,fill=white] (X1) at (2,0) {$\scriptstyle{x_1}$};
\node[point,fill=white] (X2) at (4,0) {$\scriptstyle{x_2}$};
\draw (X0) edge[color=blue!60, line width=1.4pt] node [above] {$\scriptstyle{1.4}$} (X1);
\draw (X1) edge[color=blue!60, line width=1.4pt] node [above] {$\scriptstyle{1.4}$} (X2);
\draw (X0) edge[dashed,out=315,in=225,color=red!60] node [below] {$\scriptstyle{3.5}$} (X2);
\end{scope}

\begin{scope}[shift={(7.0,0)}, scale=0.9]
\draw node at (2,1.4) {\small $\delta=0.6$};
\draw[-] (-1.5,-1.75) rectangle (5.5,1.0);
\node[point,fill=white] (X0) at (0,0) {$\scriptstyle{x_0}$};
\node[point,fill=white] (X1) at (2,0) {$\scriptstyle{x_1}$};
\node[point,fill=white] (X2) at (4,0) {$\scriptstyle{x_2}$};
\draw (X0) edge[color=blue!60] node [above] {$\scriptstyle{1.8}$} (X1);
\draw (X1) edge[color=blue!60] node [above] {$\scriptstyle{1.8}$} (X2);
\draw (X0) edge[dashed,out=315,in=225,color=red!60, line width=1.4pt] node [below] {$\scriptstyle{3}$} (X2);
\end{scope}

\begin{scope}[shift={(14.0,0)}, scale=0.9]
\draw node at (2,1.4) {\small $\delta=0.48$};
\draw[-] (-1.5,-1.75) rectangle (5.5,1.0);
\node[point,fill=white] (X0) at (0,0) {$\scriptstyle{x_0}$};
\node[point,fill=white] (X1) at (2,0) {$\scriptstyle{x_1}$};
\node[point,fill=white] (X2) at (4,0) {$\scriptstyle{x_2}$};
\draw (X0) edge[color=blue!60, line width=1.4pt] node [above] {$\scriptstyle{1.44}$} (X1);
\draw (X1) edge[color=blue!60, line width=1.4pt] node [above] {$\scriptstyle{1.44}$} (X2);
\draw (X0) edge[dashed,out=315,in=225,color=red!60] node [below] {$\scriptstyle{3.36}$} (X2);
\end{scope}

\end{tikzpicture}
\end{center}
\caption{
\small 
The graph in Figure \ref{fig:graph_quant:original} with quantized edge-weights for specific values of $\delta$ (decreasing from left-to-right). The optimal path in each subfigure is in bold.
}
\label{fig:graph_multiple}
\end{figure}

Figs. \ref{fig:graph_quant:blue} \& \ref{fig:graph_quant:red} each show the quantized cost along each path along with this upper bound. 
There are several interesting features that this example illustrates. First, the convergence of the quantized edge weights is non-monotone, as expected. Further, as $\delta \to 0$, the minimum-cost path corresponding to the quantized edge-weights switches between the two feasible paths in the graph. 
Fig. \ref{fig:graph_multiple} shows the graph with quantized weights $\myhat{c}$ for specific values of $\delta$, where the path shown in bold is the optimal quantized path. 
Finally, as $\delta$ decreases, a threshold ($\delta=0.1$) is passed where the quantized cost along the top (truly optimal) path is always less than the quantized cost along the bottom (suboptimal) path. 

\def\dir{img/s3/}

Returning to the bi-criteria planning,
the monotone decrease of errors due to quantization can be ensured if every path $P$ satisfying $\hat{\phi}(P) \leq b$ for a fixed value of $\delta$
will still satisfy the same inequality as $\delta$ decreases.  If we define a sequence $\delta_k = B/m_k$, the simplest sufficient condition is to use 
$m_{k+1} = 2 m_k$.  This is the approach employed in all experiments of Section \ref{s:bench}.
For the rest of the paper we suppress the hats on the $W$'s for the sake of notational convenience; 
the slackness measurements corresponding to each value of $\delta$ are shown for diagnostic purposes in all experiments of Section \ref{s:bench}.



\section{Benchmarking on synthetic data}
\label{s:bench}

We aim to gain further understanding of the proposed algorithm in applications of robotic path planning by focusing most of the attention on a simple scenario. 
Here, in contrast to Section 5, our goal is to study the convergence properties of the algorithm by refining two accuracy parameters: the number of PRM nodes in the graph and the number of budget levels (determined by $\delta$).  
Computations are done offline on a laptop rather than a robotic system, allowing for more computational time than in a real scenario. 
We assume that the occupancy map and threats are known a priori so that we expend no effort in learning this information. 
All tests are conducted on a laptop with an Intel i7-4712HQ CPU (2.3GHz quad-core) with 16GB of memory.

\subsection{Case Study Setup}
The results presented in this paper require a graph that accurately represents the motion of a robot in an environment with obstacles, where nodes correspond to collision-free configurations and edges correspond to feasible transitions between these configurations.  We are particularly interested in graphs that contain a large number of transitions between any two nodes.  There are a number of algorithms capable of generating such graphs, such as the Rapidly-exploring Random Graphs (RRG) \cite{karaman2011sampling} or Probabilistic Roadmap (PRM) \cite{kavraki1996probabilistic} algorithms. In this work we chose to use the PRM algorithm as implemented in Open Motion Planning Library (OMPL) \cite{sucan2012open} as a baseline algorithm to generate the roadmap/graph. 
In essence, the PRM algorithm generates a connected graph representing the robot's state (e.g. position, orientation, etc...) in the environment. The graph is generated by randomly sampling points in the state-space of the robot (called the configuration space in robotics) and trying to connect existing vertices of the graph within the sampled configuration. 

In case studies examined, we construct a roadmap as a directed graph $\Graph = (X,E)$ where $X$ is the set of nodes and $E$ is the set of transitions (edges).  
The nodes of $\Graph$ represent the positions of the robot in 2D, i.e. $\x\in X \subset \R^2$, and an edge $e = \parens{\x_i, \x_j} \in E$, $\x_i, \x_j\in X$ represents that there is a collision-free line segment from node $\x_i$ to node $\x_j$ in the PRM graph.
Each edge of $\Graph$ is assigned a primary and secondary transition cost by two cost functions $C,c:E\to\R^+$, respectively.

For the configuration space we chose $\domain \subset \R^2$ to be sampled by the OMPL planning library \cite{sucan2012open} with the default uniform (unbiased) sampling, and two slight modifications: we increased the number of points along each edge that are collision-checked for obstacles and removed self-transitions between nodes.
For real-world examples the roadmap $\Graph$ is typically generated by running the PRM algorithm for a fixed amount of time. However, for testing purposes we specify the number of nodes desired in $\Graph$.

Two test cases are presented:
\begin{itemize}
\item Sections \ref{ss:pareto_front}--\ref{ss:swap_primary_secondary} present an example that uses an occupancy map generated by the Hector SLAM algorithm \cite{kohlbrecher2014hector} via a LIDAR sensor in a real testing facility.
\item Section \ref{ss:vis_example} presents a more complicated scenario in a synthetic environment to illustrate the much broader scope of the algorithm.
\end{itemize}
Both tests assume that the physical dimensions of $\Omega$ are $450\text{m}\times450\text{m}$ (so each pixel of the occupancy map is $1\text{m}^2$).
Each edge is generated by connecting vertices in the graph that are `close together' via straight-line segments and collision-checking the transition (edge) with a sphere of radius 5m, which represents the physical size of the robot.

\begin{figure*}[h!]
	\centering
	\subfloat[]{
		\label{fig:numerical_analysis:env}%
		\includegraphics[scale=0.5]{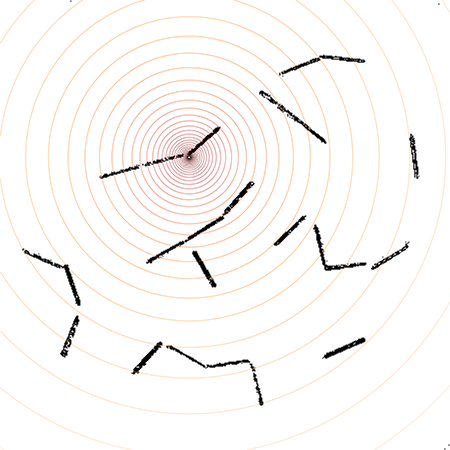}%
	}%
	\hspace{4mm}%
	\subfloat[]{%
		\label{fig:numerical_analysis:prm}%
		\includegraphics[scale=0.5]{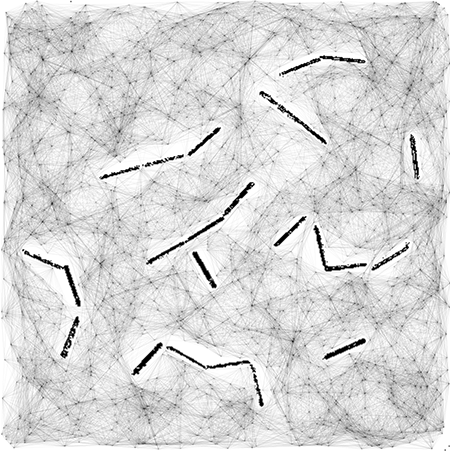}%
		}
	\caption{
		\small
		\textsc{(a)} Environment with obstacles and contours of a threat exposure function.
		\textsc{(b)} PRM Roadmap with $2,048$ nodes and $97,164$ edges.
}
\label{fig:numerical_analysis}
\end{figure*}

The first occupancy map considered is shown in Fig. \ref{fig:numerical_analysis:env}; the color black represents occupied areas (walls) whereas white represents the unoccupied areas (configuration space). 
The concentric circles represent the contours of a threat level function (to be defined).
A PRM-generated roadmap with $2,048$ nodes and $97,164$ edges over this occupancy map is shown in Fig. \ref{fig:numerical_analysis:prm}.

\def\dir{img/s4_husky/}

\subsection{Cost Functions}
\label{ss:cost_functions}

We consider two cost functions $D$ and $T$ for the multi-objective planning problem. 
The cost function $D$ represents the Euclidean distance between two adjacent nodes in $X$, i.e. $D(e) = \norm{\x_i - \x_j}$ where $e=(\x_i, \x_j)$. 

The cost function $T$ denotes the threat exposure along an edge in the graph. 
To define $T$ we assume that there are threats $\mathcal{K}=\cbraces{1, 2, \ldots ,N_T}$ in the environment, which are indicated by their position $\p_k\in \R^{2}$, severity $s_k$, minimum radius $r_k \geq 0$, and visibility radius $R_k \geq 0$ for each $k\in \mathcal{K}$. For each threat, these parameters define a threat exposure function
\begin{equation}
\label{individual_threat_equation}
\tau_{k}(\x) =
\begin{cases}
s_k / r_k^2 			& \quad \text{if } \norm{\x-\p_k} \leq r_k				\\
s_k / \norm{\x-\p_k}^2		& \quad \text{if } \norm{\x-\p_k} \in \parens{r_k, R_k}		\\
s_k / R_k^2				& \quad \text{if } \norm{\x-\p_k} \geq R_k				\\
\end{cases},
\end{equation}
for $\x\in \R^{2}$. Fig. \ref{fig:numerical_analysis:env} shows the contours of $\tau$ for a single threat with $s = 20$, $r = 5$m, and $R = +\infty$.
For simplicity we will always assume that $R_k = +\infty$ for all $k\in \mathcal{K}$ in all examples. 

The cumulative threat exposure function is then defined as the integral of the instantaneous threat exposure along the transition (edge), summed over all threats $k \in K$
\begin{equation}
\label{threat_equation}
T(e) = \sum_{k \in \mathcal{K}}  \int_0^1 \tau_k\braces{(1-t)\cdot \x_i + t \cdot \x_j} \; \norm{\x_j - \x_i} \; dt
\end{equation}
where $e=(\x_i, \x_j)\in E$ is parameterized above by assuming the robot moves at a constant speed. 
Note that in \eqref{threat_equation}, we penalize both the proximity to each threat and the duration of exposure.   
An example threat placement in the environment is shown in Fig. \ref{fig:numerical_analysis:env} along with the associated contours of the threat exposure function.

\subsection{Pareto Front}
\label{ss:pareto_front}

We use the proposed multi-objective planning algorithm to identify paths that lie on the Pareto front of the primary and secondary cost.  Our approach enables the re-use of a roadmap for searches under different objectives and constraints. 

\begin{figure}[ht]
\begin{center}
	\centering
	\hspace{4mm}%
		\rotatebox{90}{\bf\small \hspace{0.8cm}Results of b-slice MOPA}%
	\hspace{4mm}%
	\subfloat[]{%
		\label{fig:paretofront:PF}%
		\includegraphics[scale=0.35]{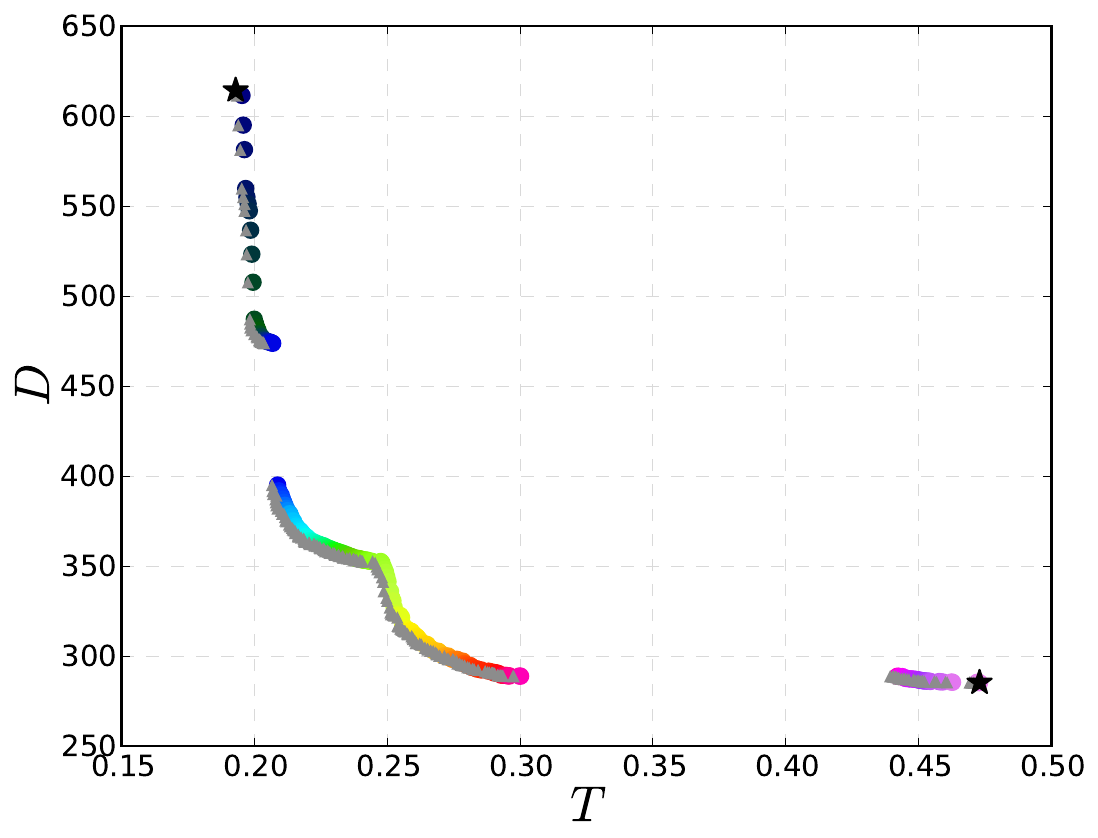}%
		}%
	\hspace{4mm}%
	\subfloat[]{%
		\label{fig:paretofront:env2}
		\includegraphics[scale=0.6]{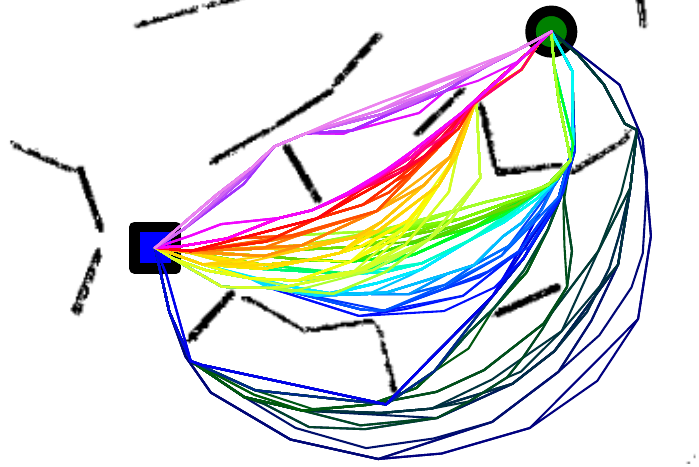}%
		}
	\\
	
		\rotatebox{90}{\bf\small \hspace{1cm}Results of scalarization}%
	\hspace{4mm}%
	\subfloat[]{%
		\label{fig:paretofront_scal:PF}%
		\includegraphics[scale=0.35]{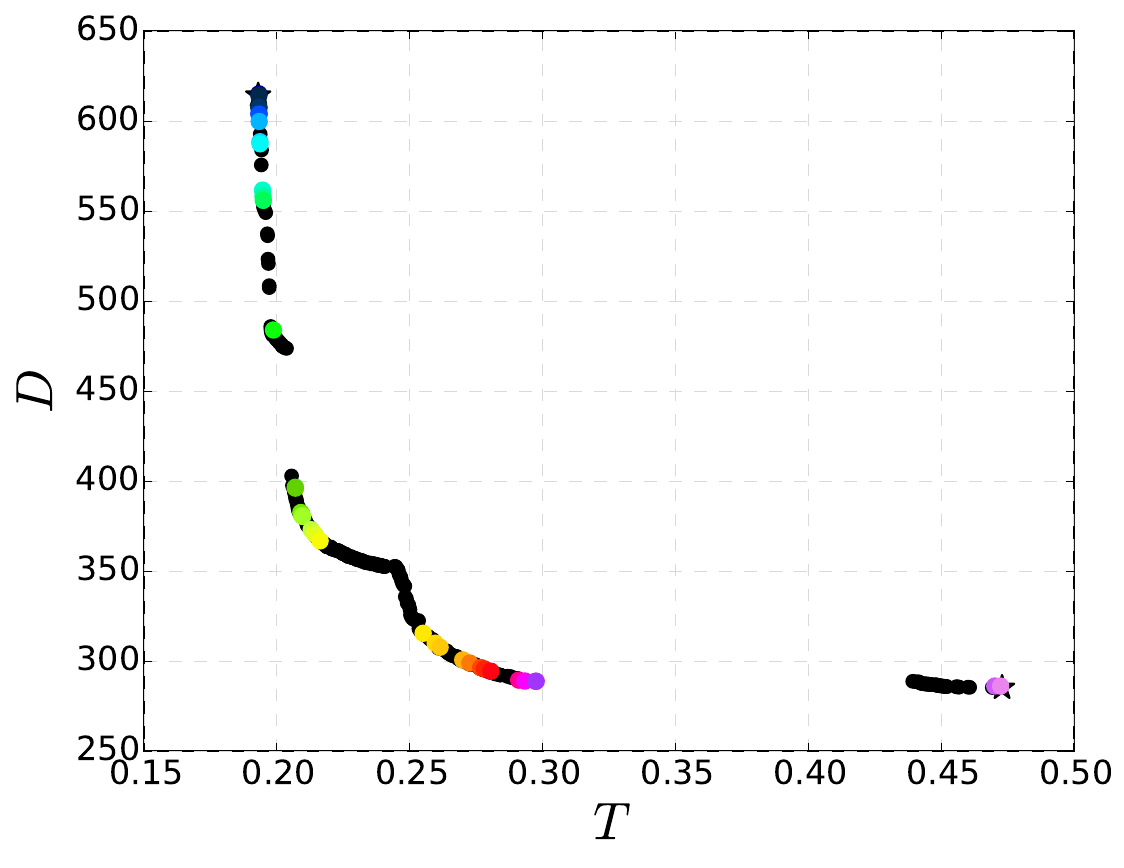}%
		}%
	\hspace{4mm}%
	\subfloat[]{%
		\label{fig:paretofront_scal:env2}
		\includegraphics[scale=0.6]{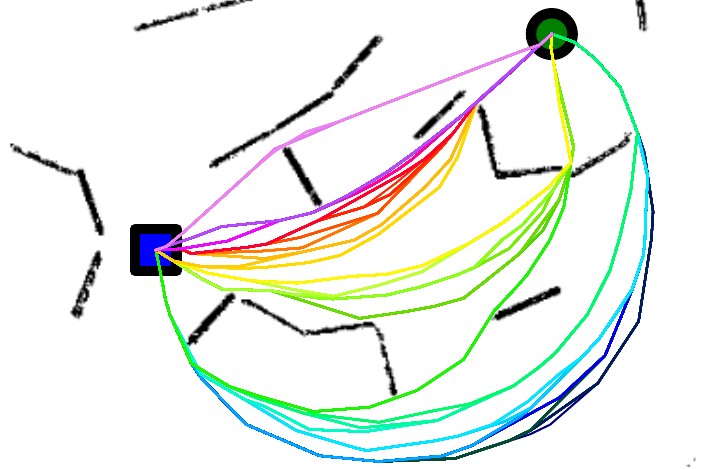}%
		}
		
	\caption{
		\small
		Test case with $C=D$ and $c=T$ and the same parameters as Figure \ref{fig:numerical_analysis}.
		\textsc{(a)} Pareto Front with a strong non-convexity. The number of budget levels was chosen to be $m = 2048$. 
		\textsc{(b)} Pareto-optimal paths in the configuration space, color-coded to correspond to the Pareto Front in Figure \ref{fig:paretofront:PF}.
		\textsc{(c)} The Pareto Front produced by scalarization algorithm is shown in color. For comparison the gray curve from Figure \ref{fig:paretofront:PF} is also shown here in black. 
		\textsc{(d)} Pareto-optimal paths in the configuration space, color-coded to correspond with the Pareto Front in Figure \ref{fig:paretofront_scal:PF}.
}
\label{fig:paretofront}
\end{center}
\end{figure}

We rely on the two cost functions $D$ and $T$ described in Section \ref{ss:cost_functions}. In every Pareto Front shown we will reserve the horizontal axis for accumulated values of $T$ and the vertical axis for accumulated values of $D$.
We first point out a Pareto Front in Fig. \ref{fig:paretofront:PF} that is generated for the environmental setup shown in Fig. \ref{fig:numerical_analysis}. The Pareto Front is color-coded (dark blue to magenta) indicating the cumulative threat exposure $T$ along the path (low-to-high), where each color-coded Pareto-optimal path within the configuration space is shown in Fig. \ref{fig:paretofront:env2}. 
As explained in Section \ref{s:implement} there is additional error due to the quantization of the secondary costs. 
As discussed, one method for measuring this error is to calculate the \textsl{true secondary cost} along each path, which is shown by the gray markers in every Pareto Front plot.

\def\dir{img/s4_husky/}

\subsection{Comparison with scalarization 
} 
\label{ss:scalarization_comparison}

The scalarization algorithm 
operates by minimizing a weighted sum of the objectives to recover Pareto-optimal solutions. 
In the context of bi-criteria path-planning on graphs, this means that we select some $\lambda \in [0,1]$  and define a new (single) cost $C_{ij}^{\lambda} = \lambda C_{ij} + (1 - \lambda) c_{ij}$ for each edge $e_{ij}.$  Dijkstra's algorithm is then applied to find an optimal path with respect to the  
new edge weights $C_{ij}^{\lambda}$'s.  Any resulting $\lambda$-optimal path is Pareto-optimal with respect to our two original criteria.  (Indeed, if some other path were {\bf strictly better} based on {\bf both} $C_{ij}$'s {\bf and} $c_{ij}$'s, it would also have a lower cumulative cost with respect to $C_{ij}^{\lambda}$'s.)  

The procedure is repeated for different values of $\lambda$ to obtain more points on $\PF$.
The computational complexity of this algorithm is $O(\ell n \log n)$, where $\ell$ is the number of sampled $\lambda$ values.
 
\begin{figure}
\begin{center}
\def\tikzPfScale{0.6}
\subfloat[]{%
\label{fig:pf_scal_tikz:1}

\begin{tikzpicture}[
point/.style={draw,shape=circle,inner sep = 0mm,minimum size = .2cm},
scale = \tikzPfScale,
>=stealth',
shorten >=1pt,
shorten <=1pt,
auto,
node distance = 1.5cm,
semithick]

\def\maxx{5}
\def\maxy{5}


\coordinate (P) at (0.39*\maxx, 0.3*\maxy);

\node[point,fill=white] (Ppt) at (P) {};
\node[below left]       (Plb) at (P) {$P$};

\coordinate (X1) at (0.05*\maxx, 1.00*\maxy);
\coordinate (Xn) at (1.00*\maxx, 0.05*\maxy);

\draw[thick]
	(X1) to[out=-85,in=175] 
	(Xn);
	
\node[point,fill=black] at (X1) {};
\node[point,fill=black] at (Xn) {};

\coordinate (H1) at (-1,0);
\coordinate (H2) at (\maxx+1,0);
\coordinate (V1) at (0,-1);
\coordinate (V2) at (0,\maxy+1);

\draw[->] (H1) -- (H2);
\draw[->] (V1) -- (V2);

\node[right] (T) at (H2) {\footnotesize $T$};
\node[below left] (D) at (V2) {\footnotesize $D$};

\coordinate (PL) at (0.39*\maxx + 1.0, 0.3*\maxy + 1.0);
\draw[->, thick] (P) to (PL);
\node[right] (PLlb) at (PL) {\scriptsize $\bm{\lambda} = (1-\lambda, \lambda)$};


  \tkzDefPoint(0.38*\maxx, 0.31*\maxy){Ppgf}
  \tkzDefPoint(0.40*\maxx, 0.29*\maxy){Rpgf}
  \tkzDefLine[parallel=through Ppgf](Ppgf,Rpgf)
  \tkzDrawLine[add = 25 and 25, color=gray, dashed](Ppgf,tkzPointResult)
  
\end{tikzpicture}
} 
%
\subfloat[]{%
\label{fig:pf_scal_tikz:2}

\begin{tikzpicture}[
point/.style={draw,shape=circle,inner sep = 0mm,minimum size = .2cm},
scale = \tikzPfScale,
>=stealth',
shorten >=1pt,
shorten <=1pt,
auto,
node distance = 1.5cm,
semithick]

\def\maxx{5}
\def\maxy{5}

\coordinate (P) at (0.25*\maxx, 0.25*\maxy);

\node[point,fill=white] (Ppt) at (P) {};
\node[below left]       (Plb) at (P) {$P$};

\coordinate (X1) at (0.15*\maxx, 1.00*\maxy);
\coordinate (X2) at (0.25*\maxx, 0.25*\maxy);
\coordinate (Xn) at (1.00*\maxx, 0.15*\maxy);

\draw[thick]
	(X1) to[out=-89,in=100] 
	(X2) to[out=-10,in=179]
	(Xn);
	
\node[point,fill=black] at (X1) {};
\node[point,fill=black] at (Xn) {};

\coordinate (H1) at (-1,0);
\coordinate (H2) at (\maxx+1,0);
\coordinate (V1) at (0,-1);
\coordinate (V2) at (0,\maxy+1);

\draw[->] (H1) -- (H2);
\draw[->] (V1) -- (V2);

\node[right] (T) at (H2) {\footnotesize $T$};
\node[below left] (D) at (V2) {\footnotesize $D$};


  \tkzDefPoint(0.24*\maxx, 0.28*\maxy){Ppgf}
  \tkzDefPoint(0.26*\maxx, 0.20*\maxy){Rpgf}
  \tkzDefLine[parallel=through Ppgf](Ppgf,Rpgf)
  \tkzDrawLine[add = 10 and 5, color=gray, dashed](Ppgf,tkzPointResult)
  
  \tkzDefPoint(0.2*\maxx, 0.26*\maxy){Ppgf2}
  \tkzDefPoint(0.32*\maxx, 0.225*\maxy){Rpgf2}
  \tkzDefLine[parallel=through Ppgf2](Ppgf2,Rpgf2)
  \tkzDrawLine[add = 3 and 6, color=gray, dashed dotted](Ppgf2,tkzPointResult)
  
  \tkzDefPoint(0.2*\maxx, 0.32*\maxy){Ppgf3}
  \tkzDefPoint(0.3*\maxx, 0.2*\maxy){Rpgf3}
  \tkzDefLine[parallel=through Ppgf3](Ppgf3,Rpgf3)
  \tkzDrawLine[add = 3.5 and 3, color=gray, dotted](Ppgf3,tkzPointResult)

\end{tikzpicture}
} 
%
\subfloat[]{%
\label{fig:pf_scal_tikz:3}

\begin{tikzpicture}[
point/.style={draw,shape=circle,inner sep = 0mm,minimum size = .2cm},
scale = \tikzPfScale,
>=stealth',
shorten >=1pt,
shorten <=1pt,
auto,
node distance = 1.5cm,
semithick,
 extended line/.style={shorten >=-#1,shorten <=-#1},
 extended line/.default=1cm,
 one end extended/.style={shorten >=-#1},
 one end extended/.default=1cm,]

\def\maxx{5}
\def\maxy{5}

\coordinate (P) at (0.125*\maxx, 0.635*\maxy);
\coordinate (R) at (0.7*\maxx, 0.16*\maxy);

\node[point,fill=white] (Ppt) at (P) {};
\node[below left]       (Plb) at (P) {$P$};
\node[point,fill=white] (Rpt) at (R) {};
\node[above right]      (Rlb) at (R) {$R$};

\coordinate (X1) at (0.05*\maxx, 1.00*\maxy);
\coordinate (X2) at (0.20*\maxx, 0.60*\maxy);
\coordinate (X3) at (0.50*\maxx, 0.55*\maxy);
\coordinate (X4) at (0.60*\maxx, 0.35*\maxy);
\coordinate (Xn) at (1.00*\maxx, 0.05*\maxy);

\draw[thick]
	(X1) to[out=-80,in=170] 
	(X2) to[out=-10,in=135] 
	(X3) to[out=-45,in=100]
	(X4) to[out=-80,in=180]
	(Xn);
	
\node[point,fill=black] at (X1) {};
\node[point,fill=black] at (Xn) {};

\coordinate (H1) at (-1,0);
\coordinate (H2) at (\maxx+1,0);
\coordinate (V1) at (0,-1);
\coordinate (V2) at (0,\maxy+1);

\draw[->] (H1) -- (H2);
\draw[->] (V1) -- (V2);

\node[right] (T) at (H2) {\footnotesize $T$};
\node[below left] (D) at (V2) {\footnotesize $D$};

\draw[extended line, dashed, gray] (P) -- (R);




  
\end{tikzpicture}
} 

\end{center}
\caption{\footnotesize 
	\textsc{(a)} Convex smooth Pareto Front 
	with a point $P$ corresponding to some specific $\lambda \in [0,1]$.
	The line perpendicular to $\bm{\lambda}$ is tangent to $\PF$ at $P$.
	If any part of $\PF$ fell below it, the path corresponding to $P$ would not be $\lambda$-optimal.
	\textsc{(b)} Convex non-smooth Pareto Front with a `kink' at the the point $P$ makes the corresponding path $\lambda$-optimal for a range of $\lambda$ values, with a different ``support hyperplane'' corresponding to each of them.  
	\textsc{(c)} Non-convex smooth Pareto Front. Points $P$ and $R$ correspond to 2 different $\lambda$-optimal paths.
	The portion of $\PF$ between $P$ and $R$ cannot be found by scalarization.
}
\label{fig:pf_scal_tikz}
\end{figure}
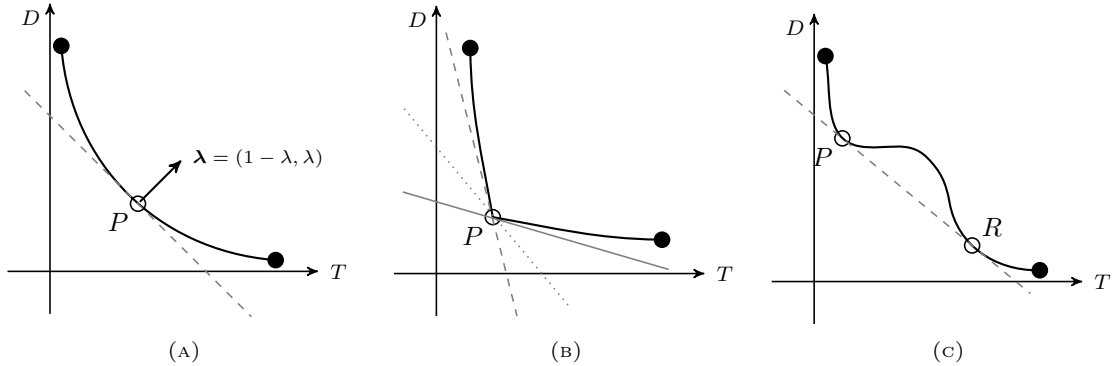

Even though scalarization recovers paths which are rigorously Pareto-optimal and does not rely on any discretization of secondary costs, it has several significant drawbacks.  First, the set of  $\lambda$ values to try is a priori unknown.    The most straightforward implementation based on a uniform 
$\lambda$-grid on $[0,1]$ turns out to be highly inefficient since the distribution of corresponding points on the Pareto front is very non-uniform.
This is easy to see geometrically, since $(\lambda-1)/\lambda$ can be interpreted as the slope of the Pareto Front wherever it is smooth; see Figure \ref{fig:pf_scal_tikz}(A).  Indeed, when $\PF$ is not smooth, we typically have infinitely many $\lambda$'s corresponding to the same point on $\PF$; see Figure \ref{fig:pf_scal_tikz}(B).
To improve the efficiency of scalarization, we have implemented a basic adaptive refinement in $\lambda$  to obtain the desired resolution of $\PF$.
The second drawback is even more significant: when the Pareto front is not convex, a single $\lambda$ value might define several $\lambda$-optimal paths, corresponding to different points on $\PF$; 
see Figure \ref{fig:pf_scal_tikz}(C).   In such cases, the Pareto Front typically is not convex, with non-convex portions remaining completely invisible when we use the scalarization.  Indeed, this approach can be viewed as approximating $\PF$ as an envelope of ``support hyperplanes'' and thus only recovers its convex hull \cite{Das}.

We tested our implementation of this adaptive scalarization algorithm 
on the example already presented in Figures \ref{fig:numerical_analysis} and \ref{fig:paretofront}. 
Not surprisingly, our budget-augmented algorithm produces significantly more (approximate) Pareto-optimal solutions than the scalarization;
see Figures \ref{fig:paretofront_scal:PF} and \ref{fig:paretofront_scal:env2}.  Regardless, of the $\lambda$-refinement threshold, scalarization completely misses all non-convex portions of $\PF$. 
In this scenario adapative scalarization produced 35 unique solutions in approximately 1.75 seconds. For comparison, our budget-augmented approach produces 160 (approximate) solutions in just 0.2 seconds.

\subsection{Effect of discretization $\delta$}

As discussed in Section \ref{s:algo}, the parameter $\delta$ determines the `discretization' of the secondary budget allowance. For all results we use $\delta = \tilde{V}_j / m$ as defined in Equation \eqref{eq:delta_M} in Section \ref{s:implement}.

\begin{figure*}[!ht]
	\centering
	\subfloat[$m=64$]{%
		\label{fig:pareto_different_levels:p0}%
		\includegraphics[scale=0.19]{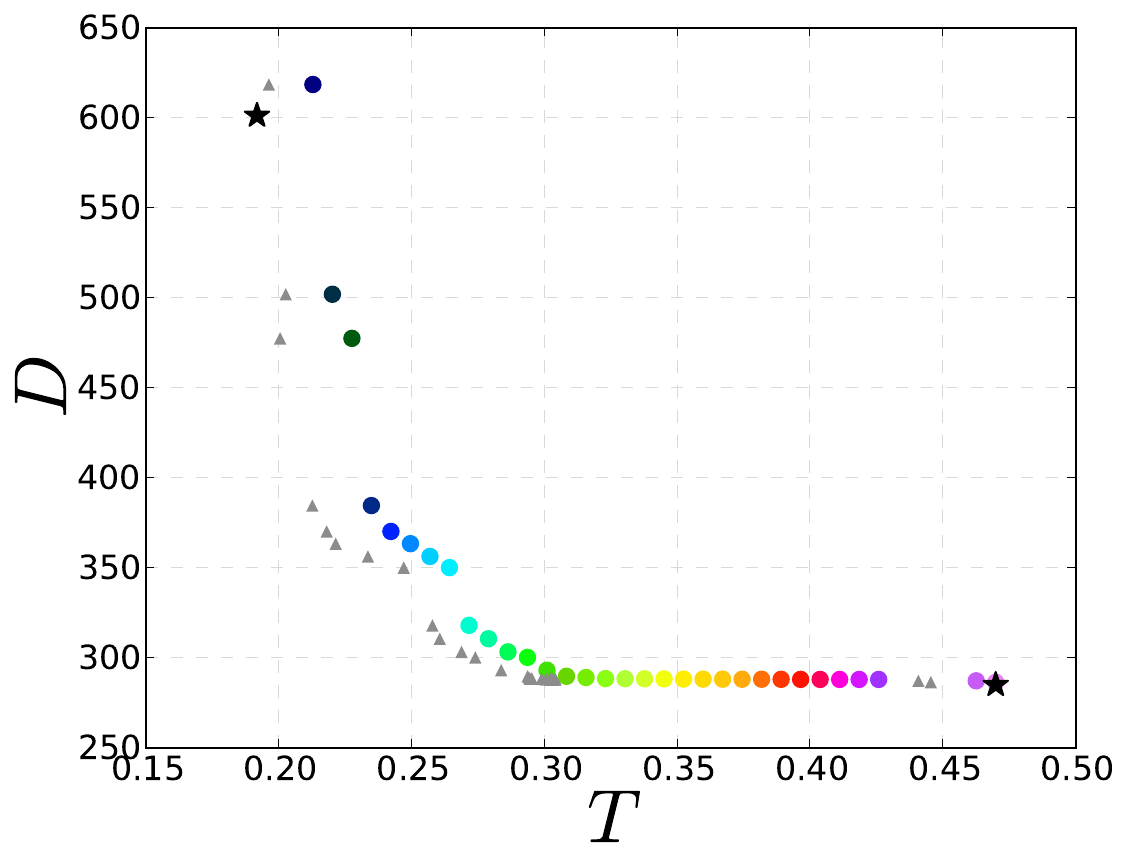}%
	}%
	\hspace{1mm}
	\subfloat[$m=128$]{%
		\label{fig:pareto_different_levels:p5}%
		\includegraphics[scale=0.19]{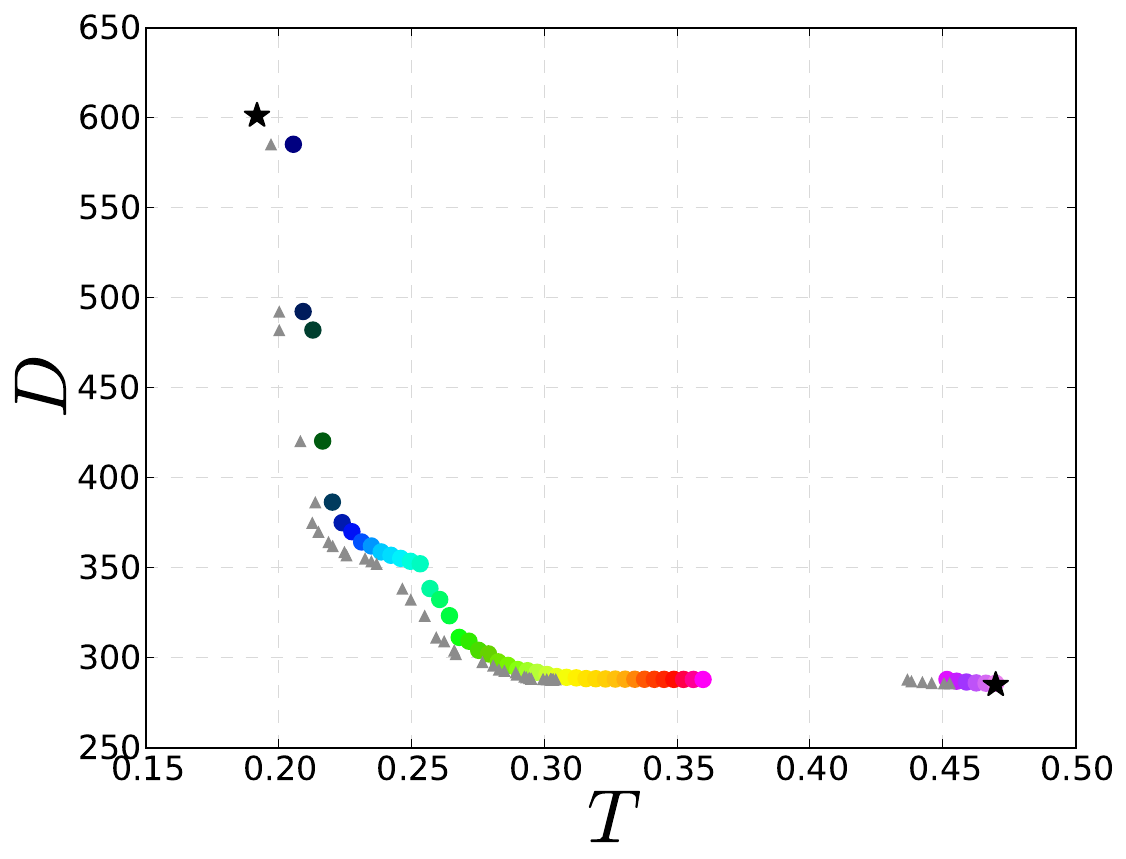}
	}%
	\hspace{1mm}
	\subfloat[$m=256$]{%
		\label{fig:pareto_different_levels:p10}%
		\includegraphics[scale=0.19]{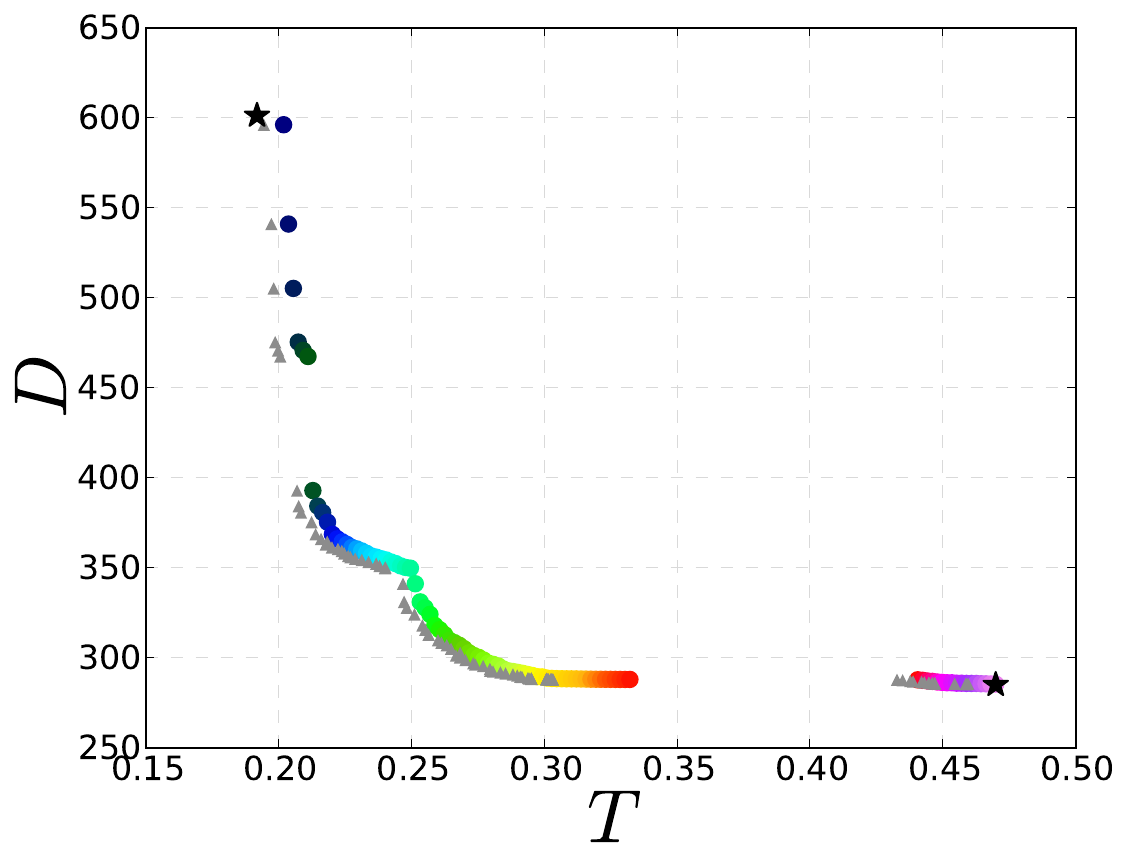}%
	}%
	\hspace{1mm}
	\subfloat[$m=512$]{%
		\label{fig:pareto_different_levels:p15}%
		\includegraphics[scale=0.19]{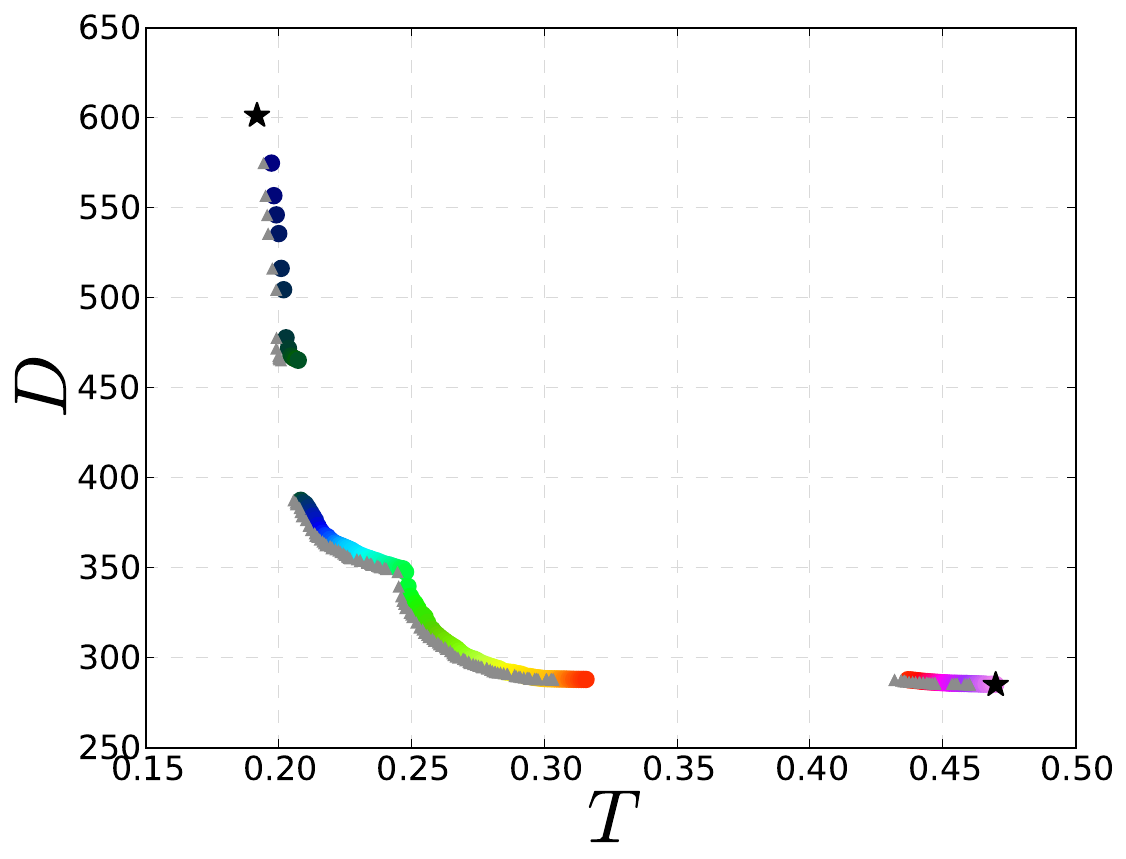}%
	} \\
	\subfloat[]{%
		\label{fig:pareto_different_levels:w0}%
		\includegraphics[scale=0.24]{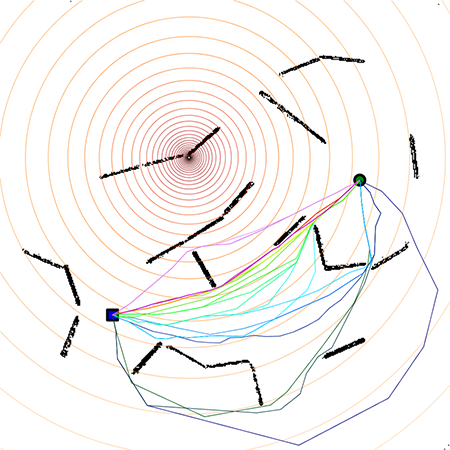}%
	}%
	\hspace{2mm}
	\subfloat[]{%
		\label{fig:pareto_different_levels:w5}%
		\includegraphics[scale=0.24]{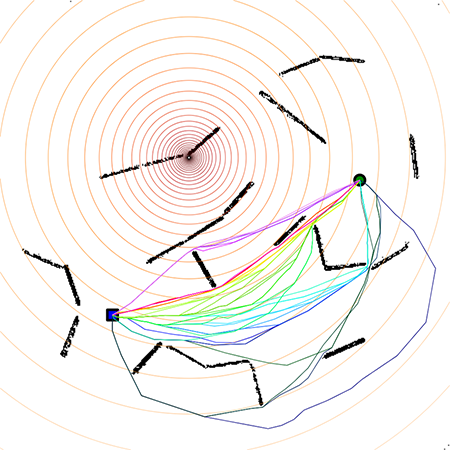}
	}%
	\hspace{2mm}
	\subfloat[]{%
		\label{fig:pareto_different_levels:w10}%
		\includegraphics[scale=0.24]{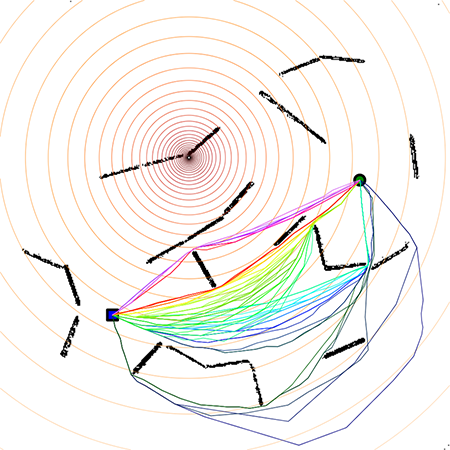}%
	}%
	\hspace{2mm}
	\subfloat[]{%
		\label{fig:pareto_different_levels:w15}%
		\includegraphics[scale=0.24]{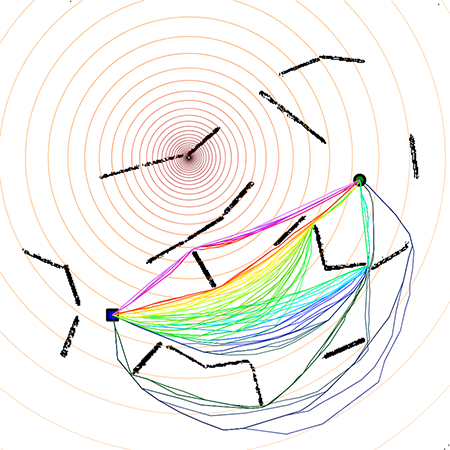}%
	} \\
	\caption{
		\small
		Test case with $C=D$ and $c=T$.
		Results correspond to a $40,000$ node graph with $3,030,612$ edges. 
		\textsc{(a)} -- \textsc{(f)}: PF and paths as $m$ (number of budget levels) varies.
		Results are qualitatively the same for $m > 512$.
}
\label{fig:pareto_different_levels}
\end{figure*}

In Fig. \ref{fig:pareto_different_levels} we generate a high-density $40,000$-node graph (with $3,030,612$ edges) and observe the effect of decreasing $\delta$ (increasing $m$) on the results. As $\delta$ decreases, it is clear that the produced Pareto Front approaches the true secondary cost curve in gray. 
The geometric sequence of $m$ values used to generate the Pareto Fronts in Figs. \ref{fig:pareto_different_levels:p0}--\ref{fig:pareto_different_levels:p15} is important as discussed in Section \ref{ss:nonmonotone_conv} -- it is what guarantees the monotone convergence. 
We note that even for non-geometric sequences of $m$ that non-monotone convergence is difficult to visually observe. 

\def\dir{img/s4_husky/}

\subsection{Swapping primary/secondary costs}
\label{ss:swap_primary_secondary}

Throughout this paper we have assumed that the roles of primary and secondary edge weights are fixed, focused on realistic planning scenarios minimizing the distance traveled subject to constraints on exposure to enemy threat(s).
However, an equivalent solution may also be obtained by switching the costs, with a 
better approximation of $\PF$ sometimes obtained without increasing the number of budget levels \cite{Rhoads}. 

\begin{figure*}[!ht]
   \centering
   \subfloat[$m=64$]{\label{fig:rev_pareto_different_levels:p0}\includegraphics[scale=0.19]{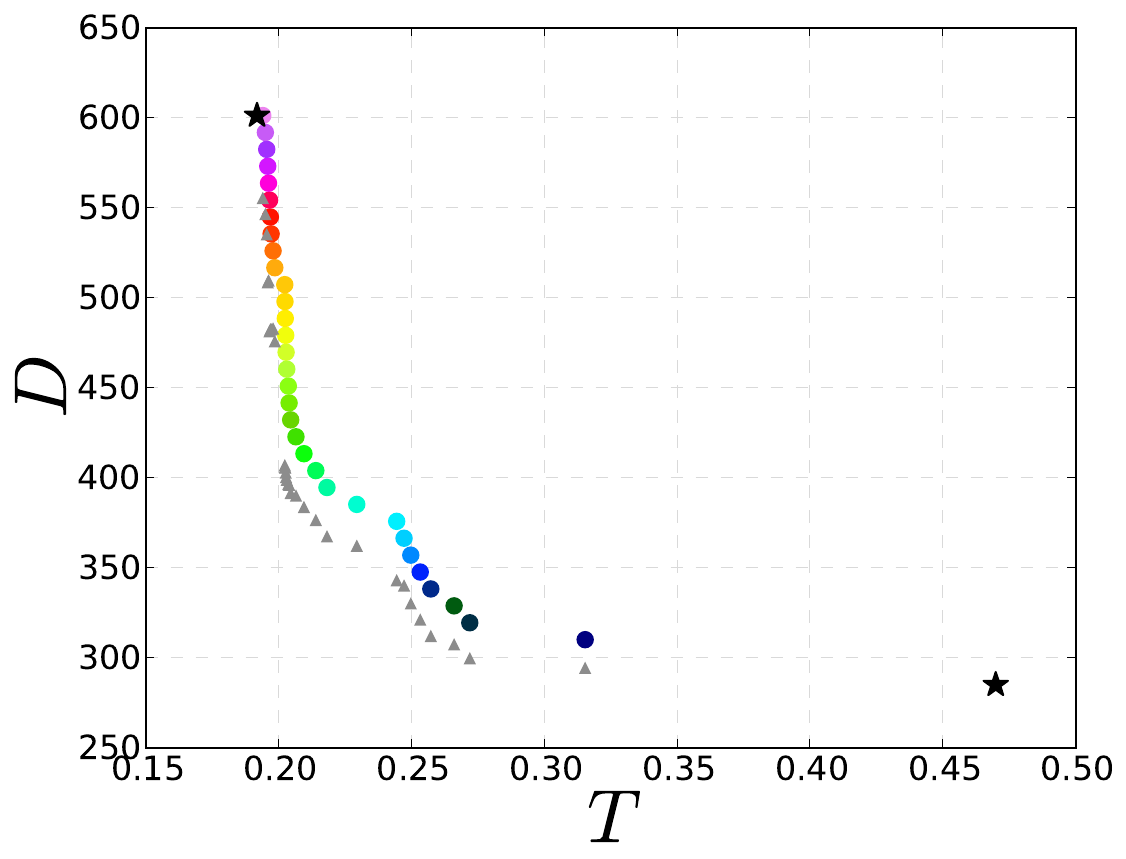}} \hspace{1mm}
   \subfloat[$m=128$]{\label{fig:rev_pareto_different_levels:p5}\includegraphics[scale=0.19]{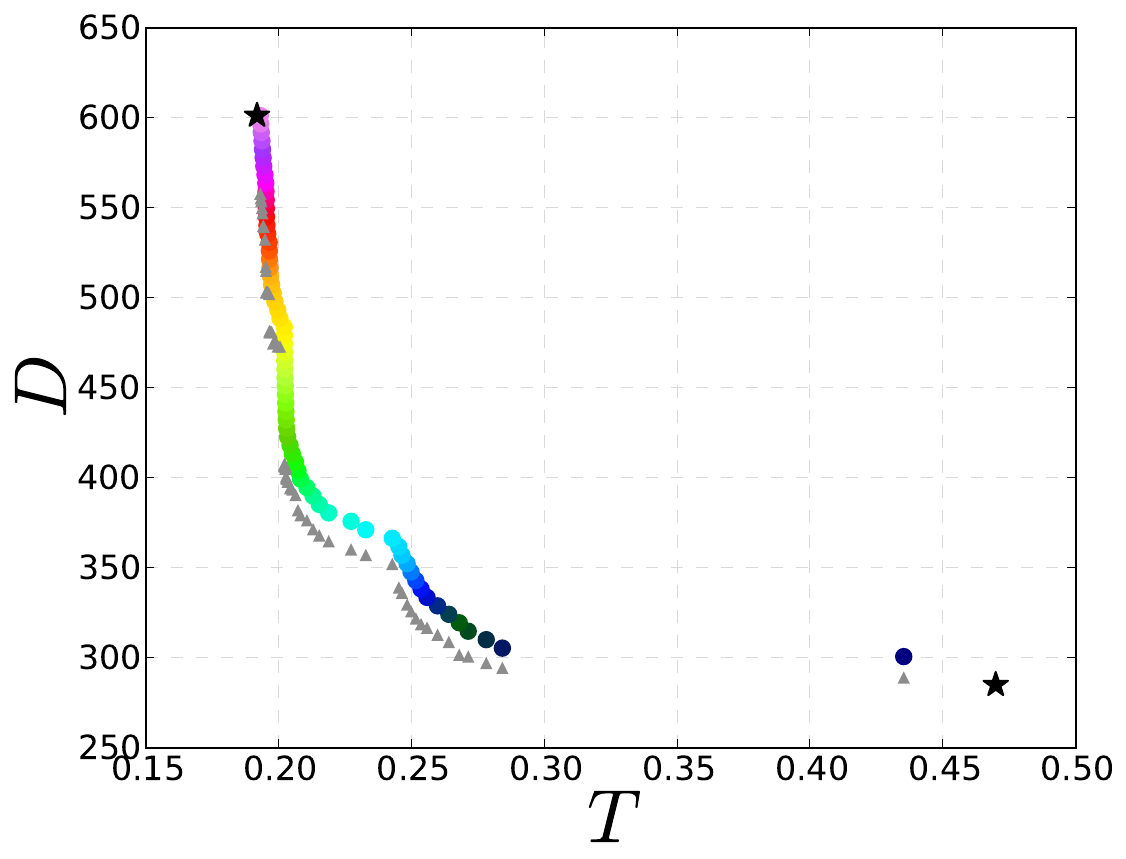}}\hspace{1mm}
   \subfloat[$m=256$]{\label{fig:rev_pareto_different_levels:p10}\includegraphics[scale=0.19]{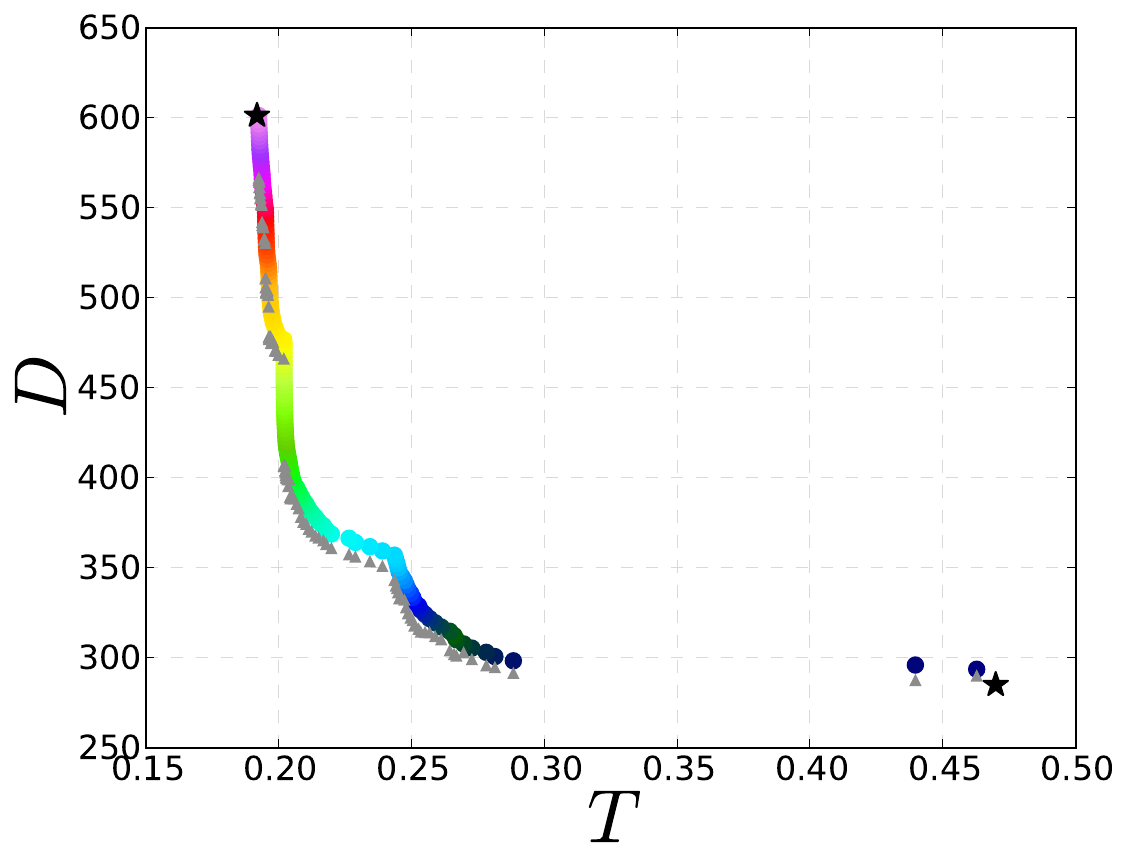}}\hspace{1mm}
   \subfloat[$m=512$]{\label{fig:rev_pareto_different_levels:p15}\includegraphics[scale=0.19]{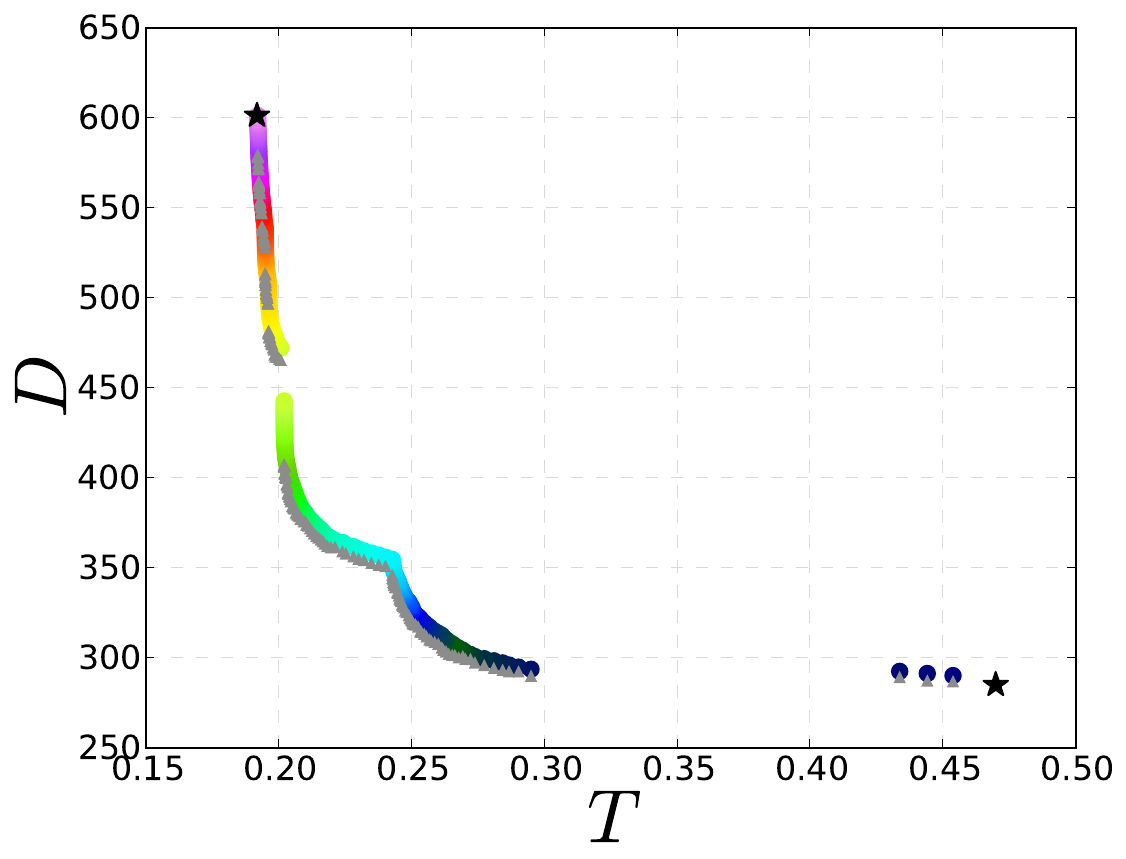}} \\
   \subfloat[]{\label{fig:rev_pareto_different_levels:w0}\includegraphics[scale=0.24]{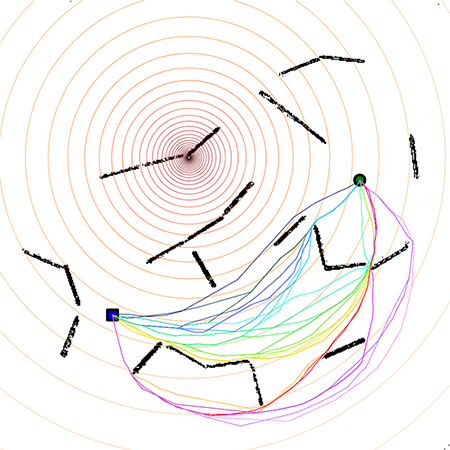}} \hspace{2mm}
   \subfloat[]{\label{fig:rev_pareto_different_levels:w5}\includegraphics[scale=0.24]{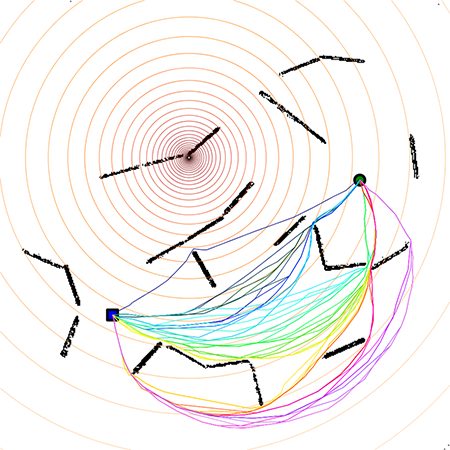}}\hspace{2mm}
   \subfloat[]{\label{fig:rev_pareto_different_levels:w10}\includegraphics[scale=0.24]{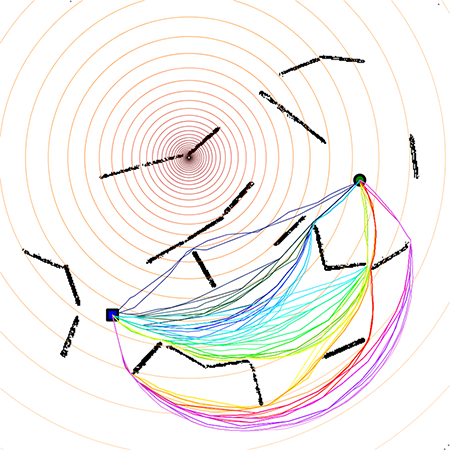}}\hspace{2mm}
   \subfloat[]{\label{fig:rev_pareto_different_levels:w15}\includegraphics[scale=0.24]{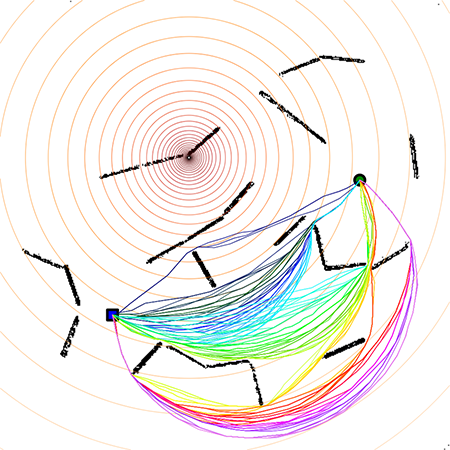}} \\
\caption{
\small
Test case with $C=T$ and $c=D$.
The same experimental setup as Fig. \ref{fig:pareto_different_levels} (except with $C$ and $c$ swapped). The Pareto Front is plotted above  the ``true secondary cost curve'' in this figure only (rather than to the right), due to the secondary objective being distance (vertical axis).
}
\label{fig:rev_pareto_different_levels}
\end{figure*}

Our experimental evidence indicates that the algorithm does a better job of recovering the portion of the Pareto Front where the slope of the front has smaller magnitude.
In other words, small changes in secondary budget result in small changes in accumulated primary cost, e.g. see Fig. \ref{fig:pareto_different_levels}.
Fig. \ref{fig:rev_pareto_different_levels} shows the result of setting the primary cost to threat exposure ($C=T$) and the secondary cost to be distance ($c=D$).
This algorithmic feature can be used to recover a more uniformly dense Pareto Front without significantly increasing the computational cost: 
consider collating the right portion of Fig. \ref{fig:pareto_different_levels:p5} and the left part of Fig. \ref{fig:rev_pareto_different_levels:p5}.

\def\dir{img/s4_husky/}

\subsection{Visibility from enemy observer}
\label{ss:vis_example}

Previous results focused on an example with a single enemy observer with a simple model for threat exposure. In this final subsection we illustrate the algorithm on a much broader example including multiple enemies with limited visibility behind obstacles.
We assume that the size of this domain $\domain$ is $450\text{m}\times450\text{m}$, and suppose there are two enemy observers with parameters:
\begin{center}
\setlength{\tabcolsep}{12pt} 
\renewcommand{\arraystretch}{1.25} 
\begin{tabular}{| c | c | c | c | c |}
\hline
\it Enemy 1	 & $s_1 = 20$	& $\p_1 = (225\text{m}, 292.5\text{m})$	& $r_1 = 5$m	& $R_1 = +\infty$	\\	\hline
\it Enemy 2	& $s_2 = 5$	& $\p_2 = (225\text{m}, 180\text{m})$		& $r_2 = 5$m	& $R_2 = +\infty$	\\	\hline
\end{tabular}
\setlength{\tabcolsep}{6pt} 
\renewcommand{\arraystretch}{1} 
\end{center}

For this problem we take the visibility of enemies into account in the construction of a new threat function $T^{vis}$. We calculate $T^{vis}$ just as in \eqref{threat_equation}, but modify the individual threat equations $\tau_k(\x)$ to account for limited visibility. 
We define modified threat level functions
\[
\tau_k^{vis}(\x)
= 
\begin{cases}
\tau_k(\x)		& \text{if the line segment from $\x$ to $\p_k$ is collision-free}	\\
\epsilon		& \text{otherwise},
\end{cases}
\]
for some small $\epsilon > 0$ (we found $\epsilon = 1/\text{Area}(\domain)$ to work well).

\begin{figure*}[!ht]
   \centering
   \subfloat[]{\label{fig:vis_example:threat}\includegraphics[scale=0.5]{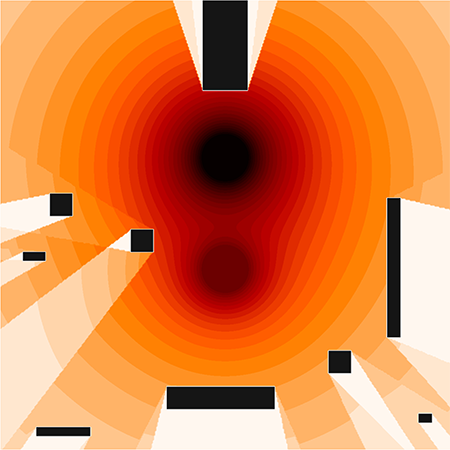}} \hspace{2mm}
   \subfloat[]{\label{fig:vis_example:roadmap}\includegraphics[scale=0.5]{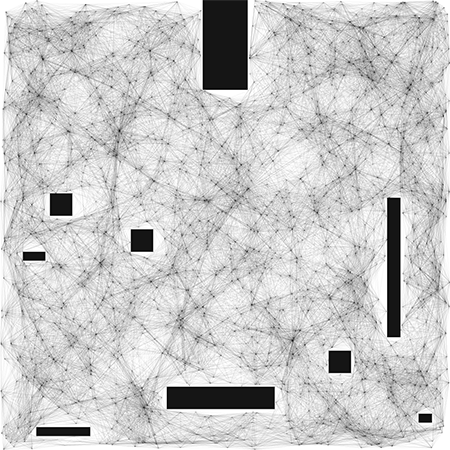}}\\
   \subfloat[]{\label{fig:vis_example:paths}\includegraphics[scale=0.44]{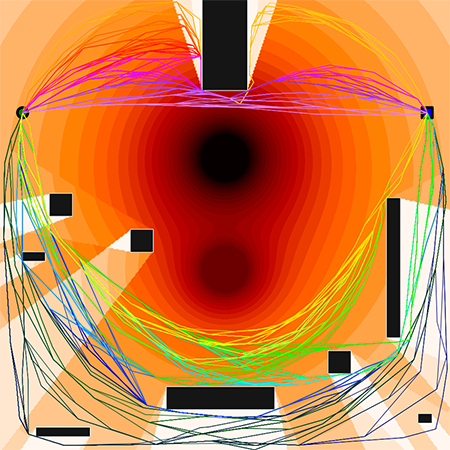}} \hspace{2mm}
   \subfloat[]{\label{fig:vis_example:pareto}\includegraphics[scale=0.45]{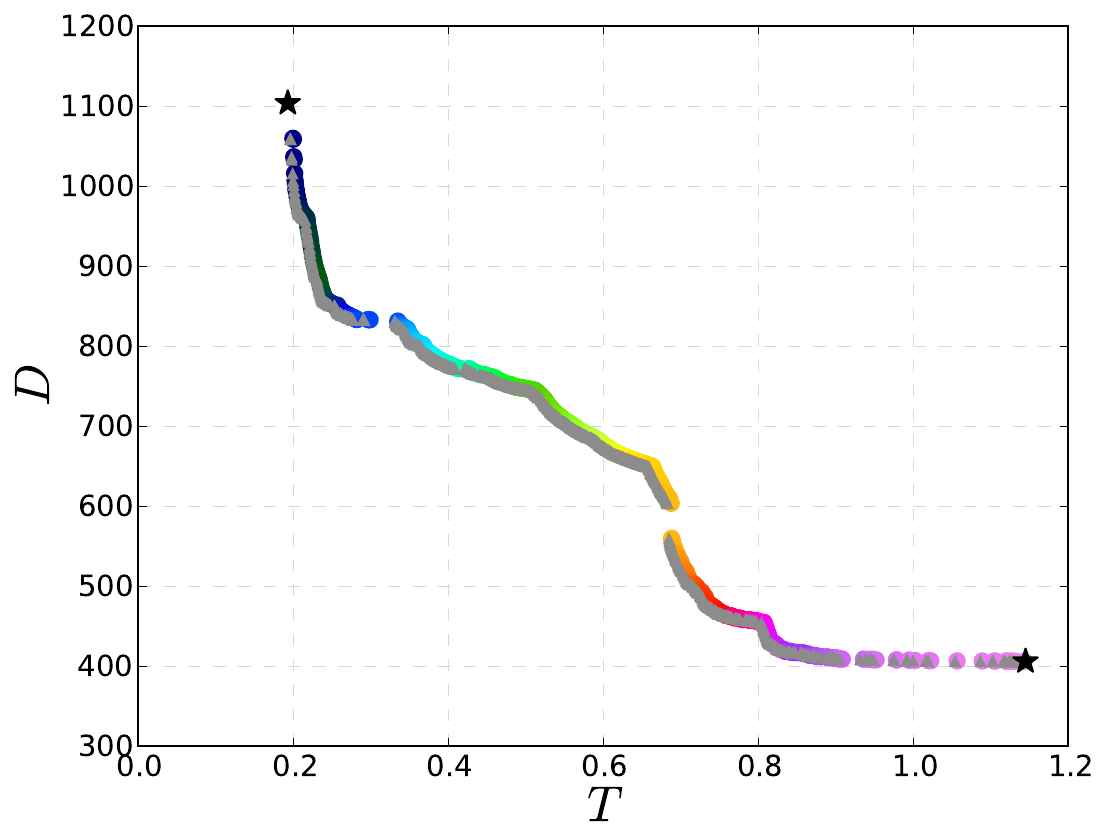}}\\
\caption{
\small
Test case with $C = D$ and $c = T^{vis}$.
\textsc{(a)} Obstacles plotted with a contour map of the threat exposure (with visibility).
\textsc{(b)} Roadmap.
\textsc{(c) \& (d)} The paths and Pareto Front corresponding to this problem setup, where $m=2,048$.
}
\label{fig:vis_example}
\end{figure*}

In Fig. \ref{fig:vis_example:threat} we show the environment where, just as before, the black rectangles represent obstacles that define the configuration space. The contours of the summed threat exposure functions, $\tau_1^{vis}(\x) + \tau_2^{vis}(\x)$, are plotted. The locations of the two enemies are visually apparent, and the visibility of each enemy can also be easily seen. In particular, the white regions in the image show the points $\x \in \Omega$ where visibility is obstructed for both enemies, i.e. $\tau_k^{vis}(\x) \equiv \epsilon$ for all $k=1,2$.
Fig. \ref{fig:vis_example:roadmap} shows the roadmap generated for a $2,048$ node graph with $103,672$ edges in this environment.

Figs. \ref{fig:vis_example:paths} \& \ref{fig:vis_example:pareto} show the results with the settings $C = D$ and $c = T^{vis}$. 
Fig. \ref{fig:vis_example:paths} shows the results of the algorithm using the same roadmap (Fig. \ref{fig:vis_example:roadmap}) with $m=2,048$.
The corresponding Pareto Front in Fig. \ref{fig:vis_example:pareto} exhibits strong non-convexity on this particular domain and generated graph due to the large number of obstacles.

\begin{figure*}[!ht]
   \centering
   \includegraphics[scale=0.65]{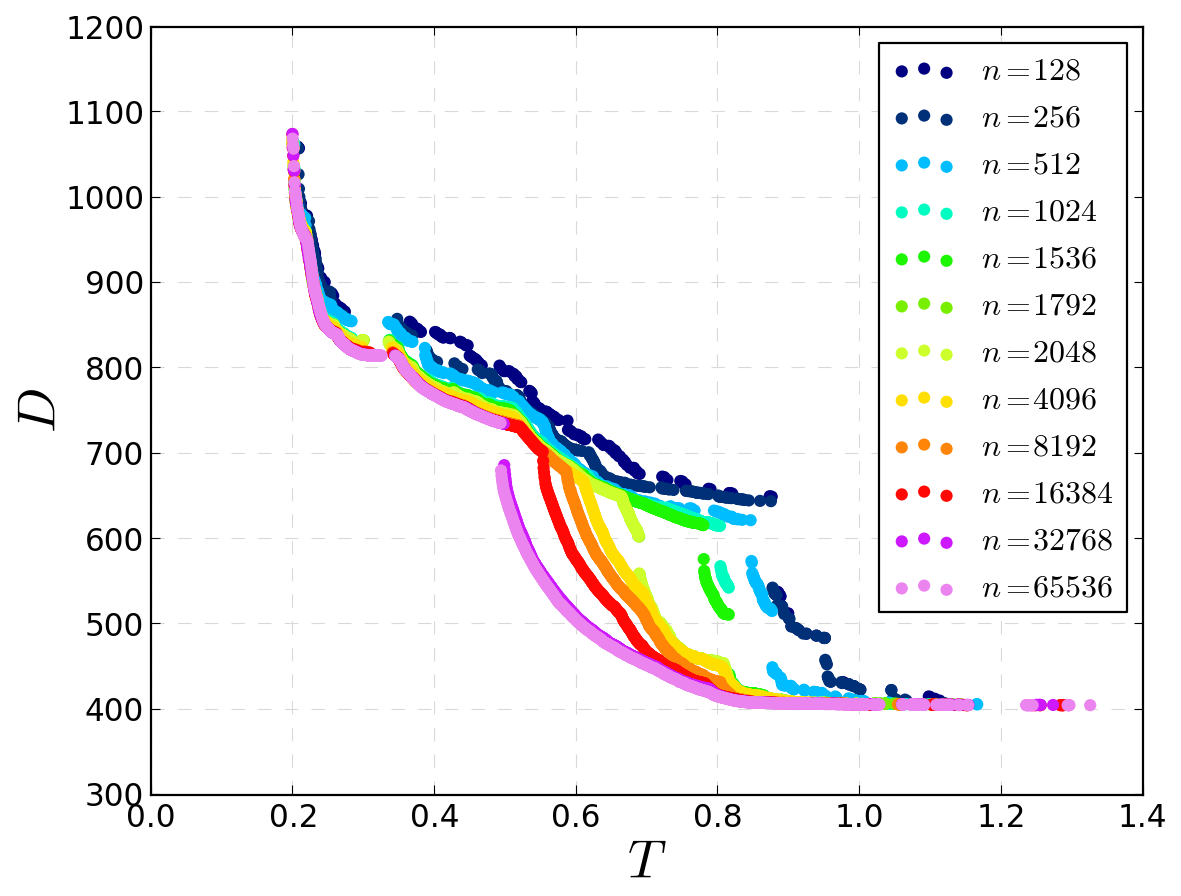}\\
\caption{
\small
Test case with $C=D$ and $c=T^{vis}$.
Using the same setup as Fig. \ref{fig:vis_example}, we study the effect of increasing the number of nodes $n$ in the graph.
The budget levels were finely discretized using $m=2048$.
}
\label{fig:vis_example_PF_test}
\end{figure*}

Our final test is to use this visibility example to study the convergence of the Pareto Front as the number of nodes in the graph increases. 
We fix $m=2048$ budget-levels in order to accurately capture the Pareto Front $\text{PF}_n$ corresponding to the generated graph of $n$ nodes. We plot $\text{PF}_n$ in Fig. \ref{fig:vis_example_PF_test} as $n$ varies.
There is some Pareto Front PF corresponding to a `continuous' version of this multi-objective problem, and we see that as $n$ increases $\text{PF}_n$ converges to PF.

\section{Experimental data}
\label{s:experiment}

\begin{figure}[th!]
	\centering
	\subfloat[]{\label{fig:husky}\includegraphics[width=0.45\hsize]{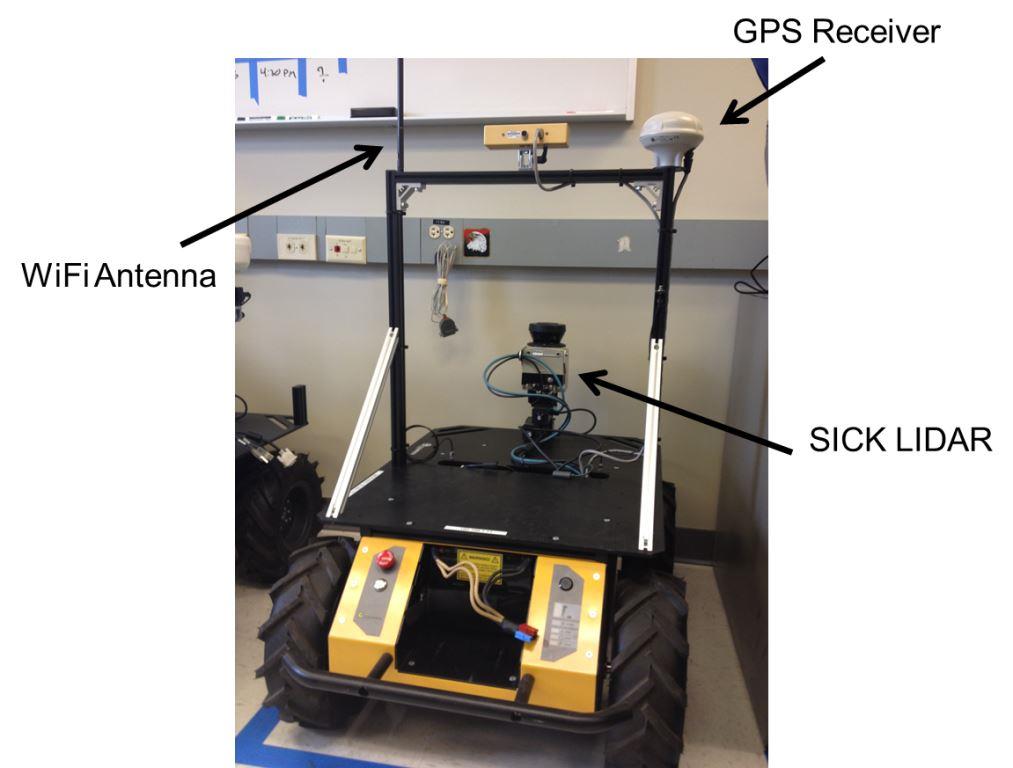}}\hfill
	\subfloat[]{\label{fig:obstacles}\includegraphics[width=0.45\hsize]{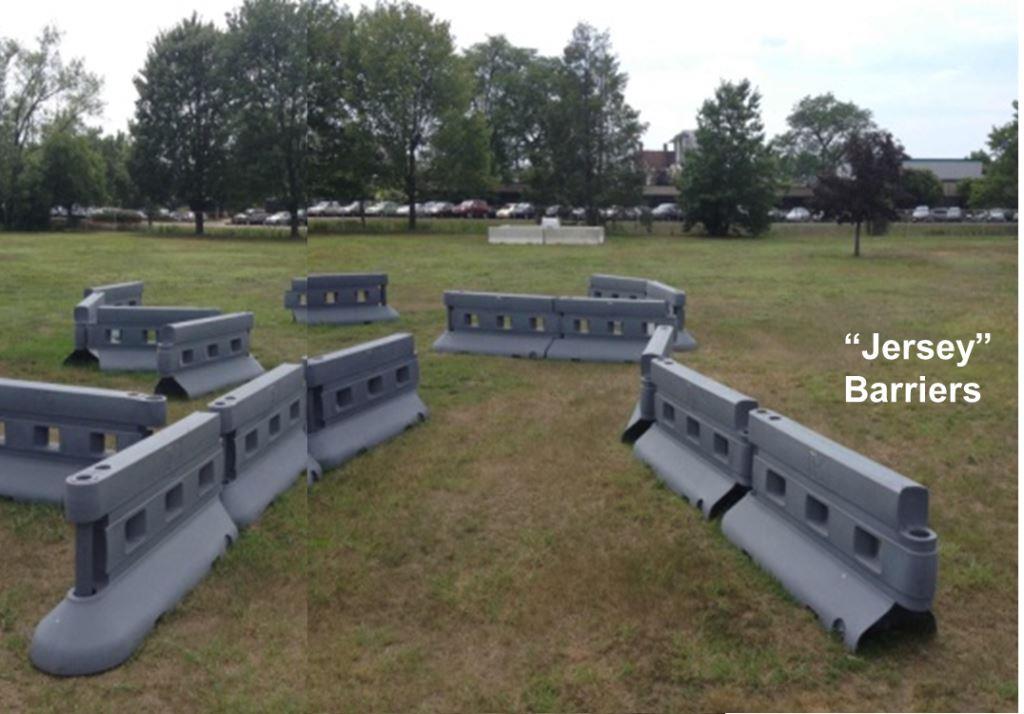}}
	\subfloat[]{\label{fig:slam}\includegraphics[width=0.45\hsize]{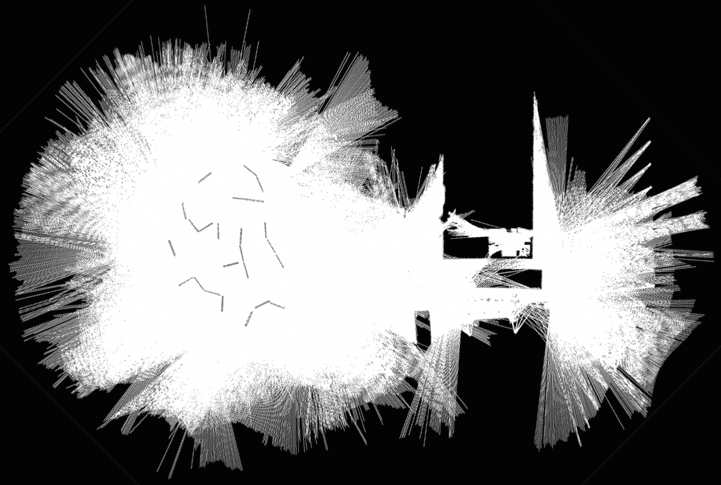}}
	\caption{
	\small
	\textsc{(a)} Husky robot from ClearPath Robotics. The GPS unit, WiFi antenna and SICK planar LIDAR are shown. A camera is also visible on the roll bar, but is was not used in the experiments.
	\textsc{(b)} Complex obstacle course created with ``Jersey barriers''. 
	\textsc{(c)} Map generated by Hector SLAM for the obstacle course.}
	\label{fig:robot_obstacle}
\end{figure}

The multi-objective path planning algorithm described in this paper was implemented and demonstrated on a Husky ground vehicle produced by Clearpath Robotics~\cite{clearpath_husky}. A picture of the vehicle used in the experiments is shown in Figure~\ref{fig:husky}. The vehicle was equipped with a GPS receiver unit, a WiFi interface, a SICK LMS111 Outdoor 2D LIDAR~\cite{sick_lidar}, an IMU, wheel encoders and a mini-ITX single board computing system with a 2.4GHz Intel i5-520M processor and 8GB of RAM. 

Although the GPS unit is used for localization, the associated positional uncertainty typically cannot meet requirements needed for path planning.  Moreover, a map of the environment is required to generate a roadmap to support path planning as discussed in the previous section.  To this end, we use a Simultaneous Localization and Mapping (SLAM) algorithm for both localization and mapping, and in particular, the Hector SLAM open source implementation \cite{kohlbrecher2014hector}.
Hector SLAM is based on scan matching of range data which is suitable for a 2D LIDAR such as the SICK LMS111.
The output of Hector SLAM is an occupancy grid representing the environment and a pose estimate of the vehicle in the map.  Details of the algorithm and implementation can be found in reference \cite{kohlbrecher2014hector} and the accuracy of localization is addressed in \cite{SantosSlamLocalizationError}.
In order to allow stable and reliable state estimation, we also implemented an extended Kalman filter (EKF). Position estimates were computed by fusing inertial information and SLAM pose estimates through the EKF. 

A set of 24 ``Jersey barriers'' placed on flat ground were used to create a complex maze-like environment with multiple interconnecting corridors. Figure~\ref{fig:obstacles} shows the barriers and their positions on the ground.  The map generated by Hector SLAM for the obstacle course is shown in Figure~\ref{fig:slam}. The map of the environment is known to the vehicle before the start of the multi-objective planning mission.
We utilized the Open Motion Planning Library (OMPL) \cite{SucanOMPL} which contains implementations of numerous sampling-based motion planning algorithms. 
We have prior experience using this library for sampling-based planner navigation of an obstacle-rich environment \cite{Kannan}. 
The generality of OMPL facilitated the development and implementation of our field-capable bi-criteria path planning algorithm: efficient low-level data structures and subroutines are leveraged by OMPL (such as the Boost Graph Library (BGL) \cite{SiekBoost}) and several predefined robot types (and associated configuration spaces) are available. For the purposes of Husky path planning we apply the PRM algorithm within an $\mathbb R^{2}$ state space.

\begin{figure}[th!]
	\centering
	\subfloat[]{\label{fig:set_goal}\includegraphics[width=0.485\hsize]{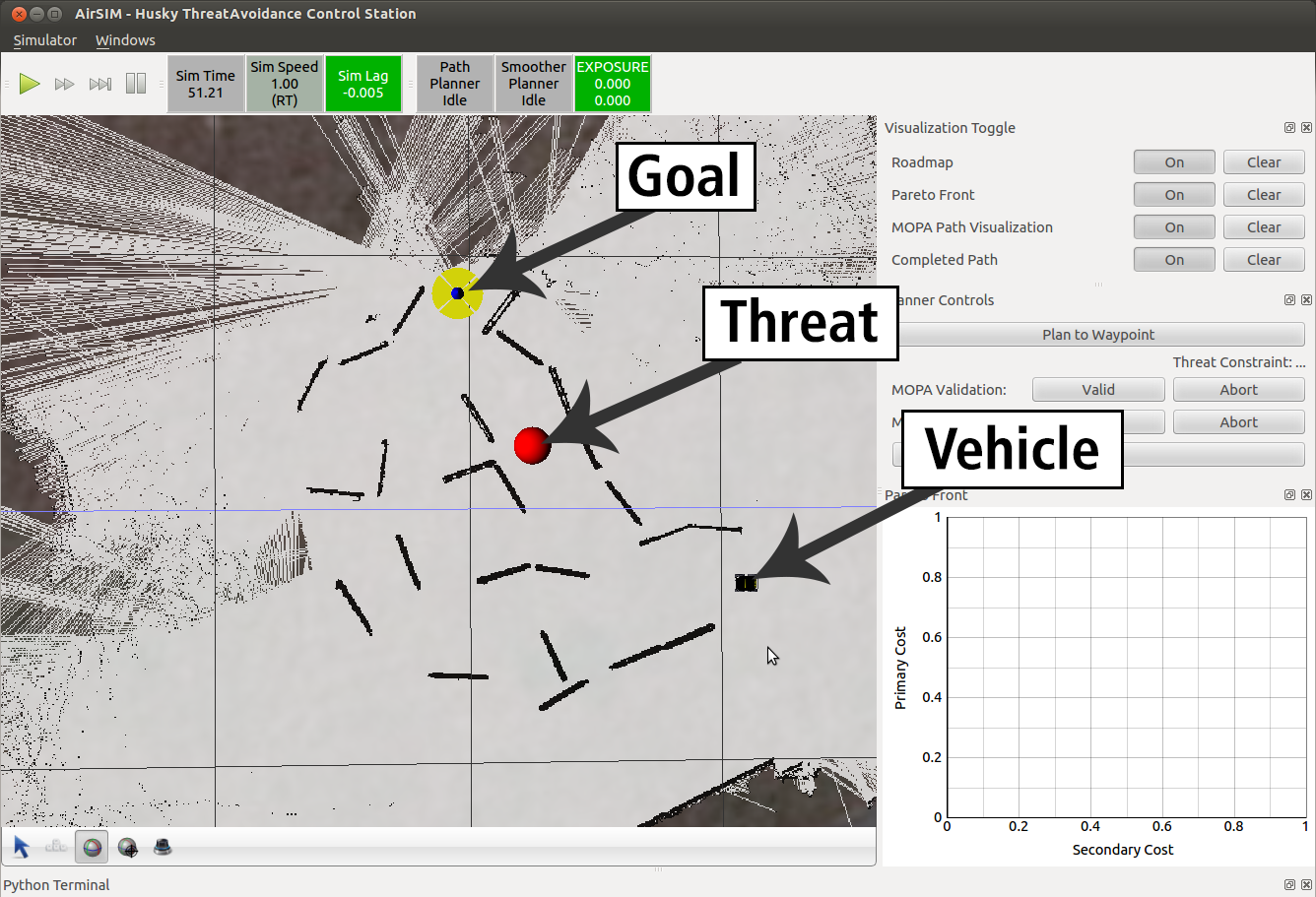}}\hfill
	\subfloat[]{\label{fig:roadmap_pf}\includegraphics[width=0.485\hsize]{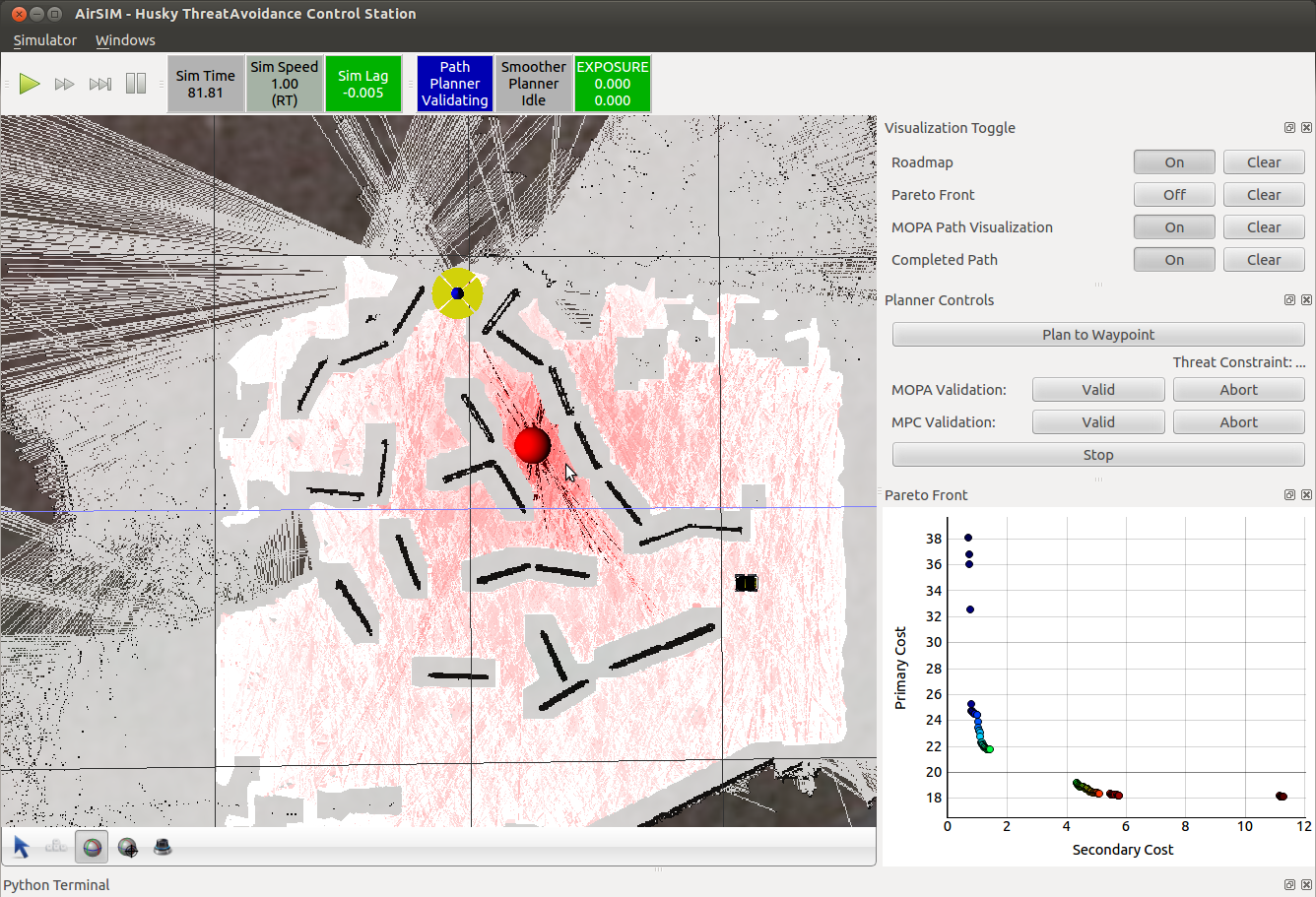}}\\
	\subfloat[]{\label{fig:click_pf}\includegraphics[width=0.485\hsize]{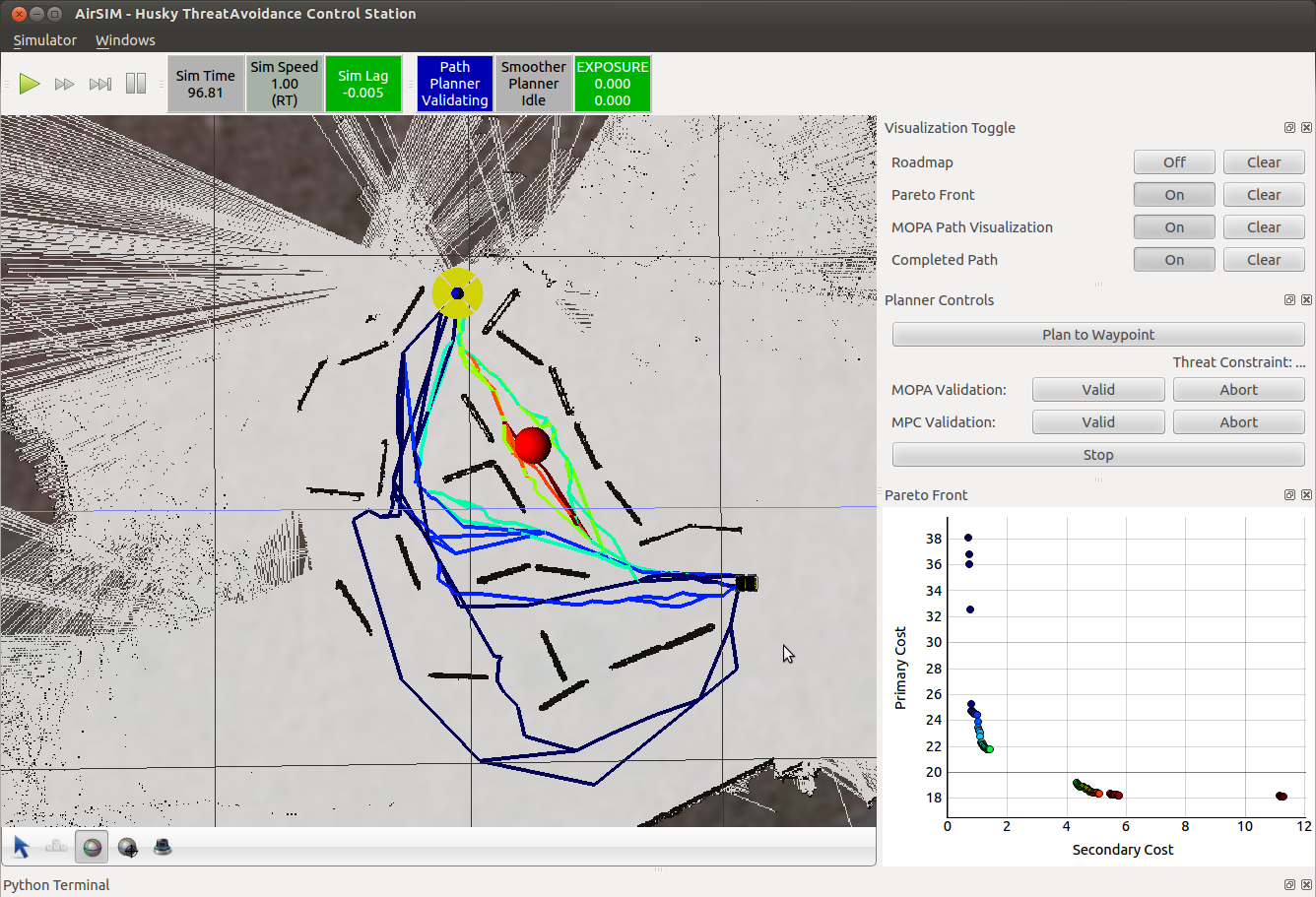}}\hfill
	\subfloat[]{\label{fig:execute_path}\includegraphics[width=0.485\hsize]{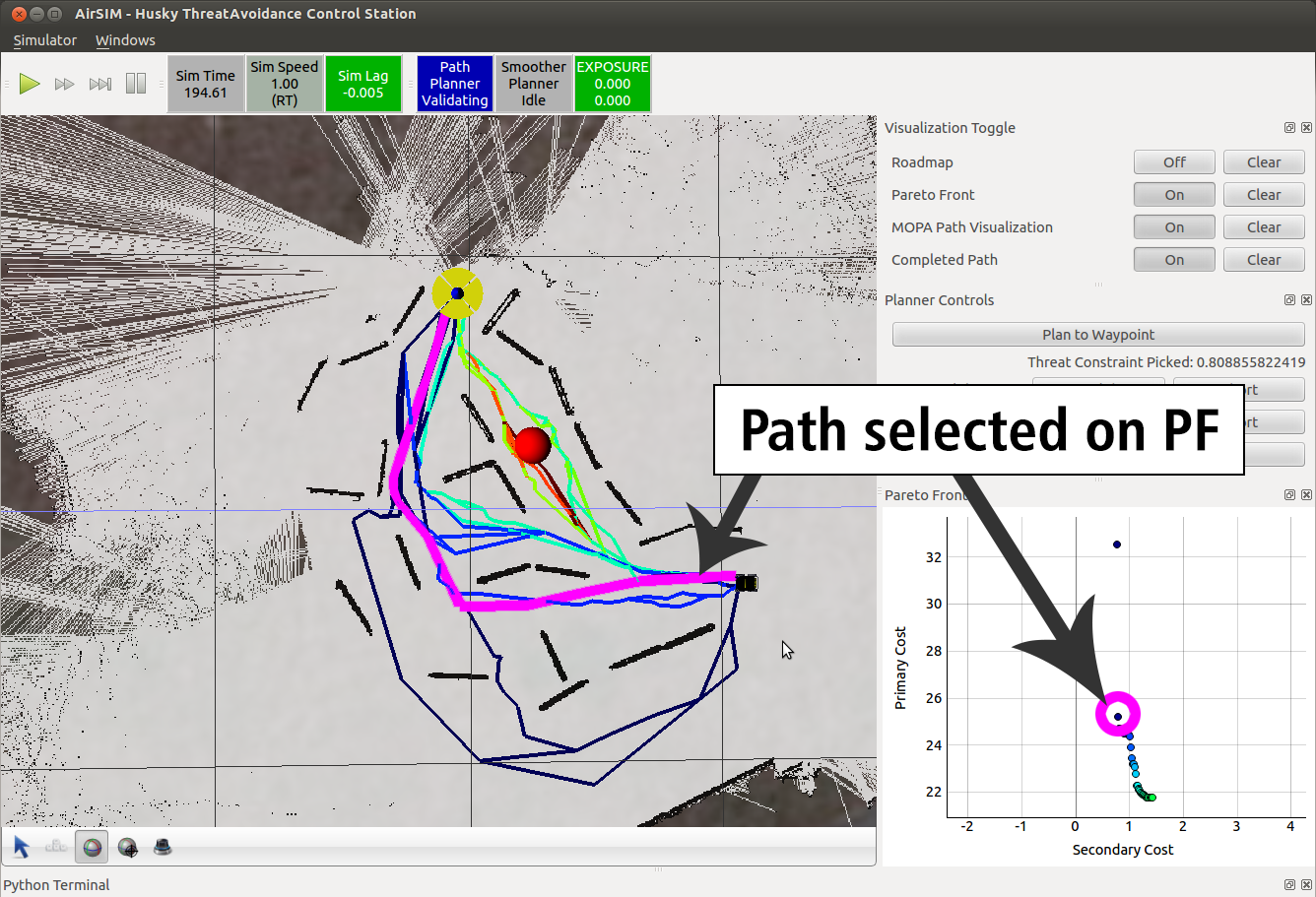}}
	\caption{
	\small
	\textsc{(a)} The user sets the goal by moving the marker to a location in the occupancy map. The vehicle and threat locations are shown.
	\textsc{(b)} A roadmap is generated with edges colored according to the secondary cost (threat exposure). The algorithm is run and the Pareto Front is visualized where the vertical and horizontal axes and primary and secondary costs, respectively.
	\textsc{(c)} The Pareto-optimal paths are shown, each corresponding to a dot on the Pareto Front (by color association).
	\textsc{(d)} The expert user clicks on a dot in the plot in the lower right corner which highlights the corresponding path in the GUI, choosing a path based on the desired trade-off between primary and secondary cost. The Pareto Front has been zoomed-in for this subfigure.
	}
	\label{fig:planning}
\end{figure}

For the experiments provided in this section, we set the number of budget levels to $m=768$.  The roadmap ``grow-time'' is set to be about $0.25$ seconds, and planning boundary to be a box of size $25m \times 25m$, resulting a roadmap with about $n = 8,000$ vertices and $116,000$ edges (e.g., see Fig. \ref{fig:roadmap_pf}).  The time to generate the roadmap graph and assign edge-weights is about $1$ second, and time to compute the solution and generate the Pareto Front is about $1$ second. Thus, a fresh planning instance from graph generation to computing a solution takes about $2$ seconds.  We also implemented a number of features in the planner intended to add robustness and address contingencies that may arise during the mission, such as when threat values in the environment change.  In this case, the execution framework will attempt to re-plan with updated threat values.  Since the graph need not be re-generated, replanning is typically much faster (about $1$ second).

In the subsequent mission, we assume the Husky is tasked with navigating from its current location to a designated goal waypoint, while avoiding threat exposure to a set of fixed and given threats, i.e., we aim to minimize distance (primary cost) subject to a maximum allowable level of threat exposure (secondary cost).  We use the cost functions $D$ (distance) and $T$ (threat exposure) as given in Section 4.2.  For threat exposure, we use a single threat with $s=1$, $r=0$ and $R=+\infty$.  Once the path is computed, the path is sent down to the lower level path smoothing module.  The path smoother is part of the hierarchical planning and execution framework as discussed in \cite{ding2014hierarchical} and is implemented using model predictive control with realistic vehicle dynamics.

Since threat exposures are difficult to quantify absolutely but easy to compare relatively, we designed a Graphical User Interface (GUI) to pick a suitable path from the Pareto front reconstructed by the proposed algorithm, i.e., the user chooses the shortest path subject to an acceptable amount risk from among the paths in the Pareto front.
The process to execute the planning mission is illustrated in Fig. \ref{fig:planning} can be described as follows:
\begin{enumerate}[1.]
    \item 
    Given an occupancy map of the environment generated by the Hector SLAM algorithm 
    and the known threat location, the user sets the desired goal location for the mission 
    (see Fig.~\ref{fig:set_goal}). 
    The user has some predefined level of tolerance/budget for the amount of threat 
    exposure allowed. Say, for example, a threat tolerance of $0.79$ is permitted in the example shown in Figure \ref{fig:planning} (for reference, the shortest distance path has a threat exposure of $11$).
    \item 
    An initial PRM is generated with edge-weights computed by primary and secondary cost functions.
    Algorithm \ref{alg:DSP_onesweep} is executed and the results are used to compute the 
    Pareto Front (see Fig. \ref{fig:roadmap_pf}).
    Next, the path corresponding to each point on the PF is recovered and shown in the GUI, 
    where color indicates the amount of secondary cost (see Fig. \ref{fig:click_pf}).  
    \item 
    From the paths available in Fig. \ref{fig:click_pf} and the amount of threat exposure tolerated, an expert analyzes the results. Originally the budget was defined to be $0.79$, which corresponds to a path with a length of $32.5$m (primary cost). The expert realizes if the exposure allowance is slightly increased to $0.81$, a path with length $25.25$m is available. These two points correspond to the large vertical drop in the Pareto Front in Fig. \ref{fig:planning}.
    \item 
    From the results and analysis, the user then chooses a path by clicking on a dot in the lower right plot.  
    The picked path is highlighted in bold (see Fig.~\ref{fig:click_pf}).
    The user then clicks a button in the GUI to verify the path as desired (see Fig. \ref{fig:execute_path}), and finally the selected path is sent to lower level modules for path smoothing and vehicle control.
\end{enumerate}

\begin{figure}[b!]
	\centering
	\subfloat[]{\label{fig:t0}\includegraphics[width=0.485\hsize]{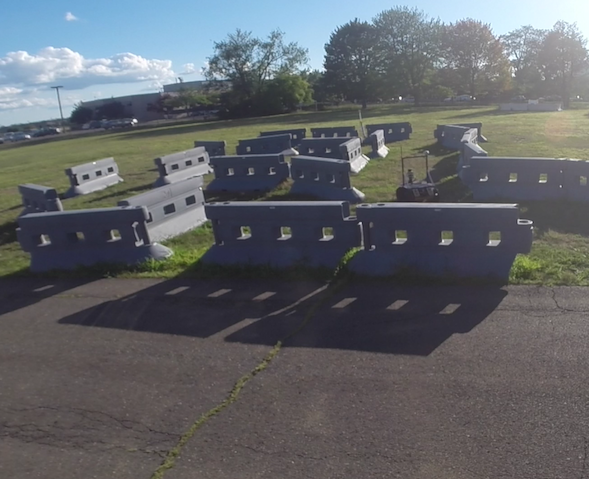}}\hfill
	\subfloat[]{\label{fig:t1}\includegraphics[width=0.485\hsize]{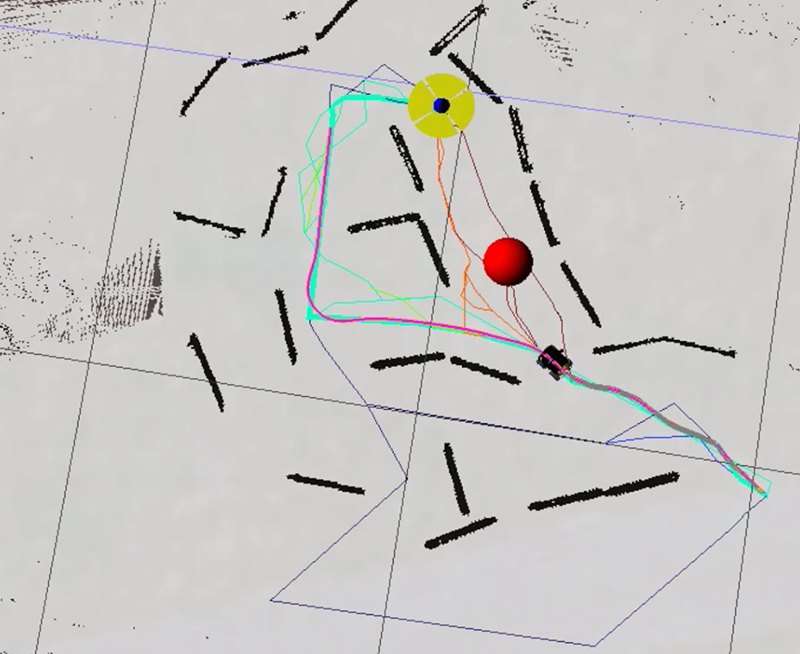}}\\
	\subfloat[]{\label{fig:t0}\includegraphics[width=0.485\hsize]{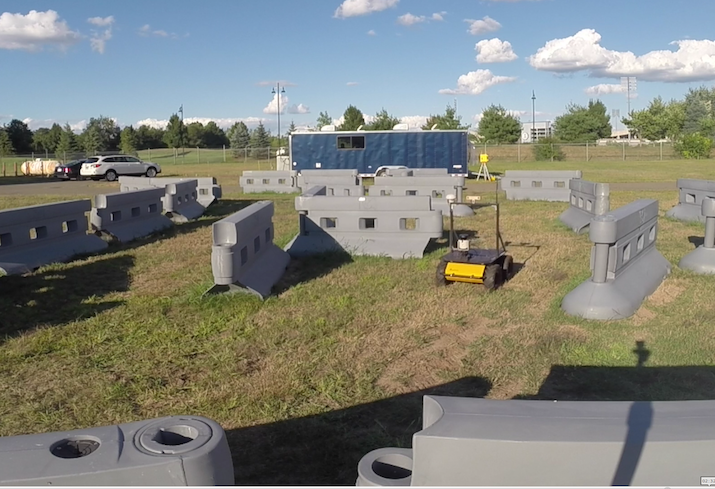}}\hfill
	\subfloat[]{\label{fig:t1}\includegraphics[width=0.485\hsize]{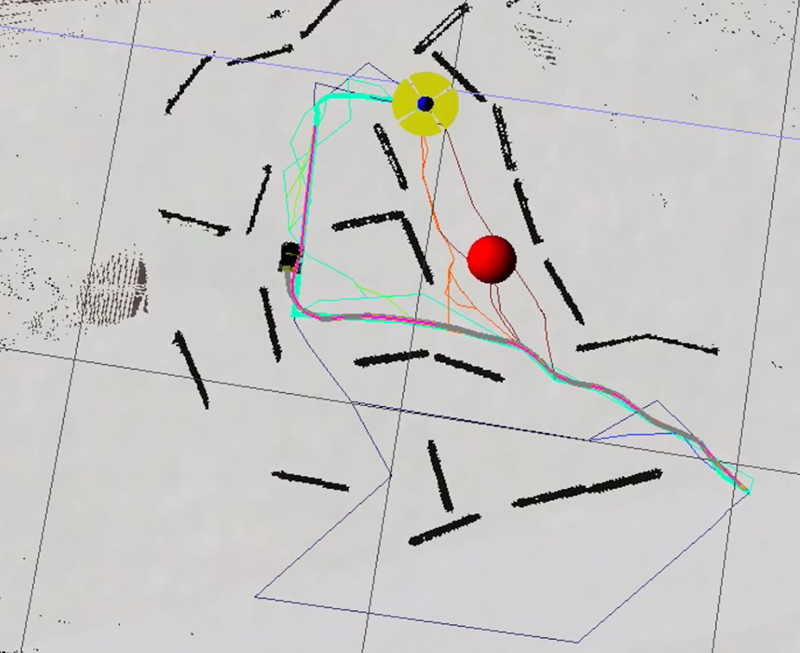}}
	\caption{
	\small
	At left, images of the experiment; at right, details of the monitoring station showing the status of the vehicle (the black rectangle), the threat location (the red dot), and the optimal trajectories computed by the algorithm.  
	\textsc{(a)} and \textsc{(b)} are the vehicle at near start of the mission and \textsc{(c)} and \textsc{(d)} are the vehicle near the end of the mission.
	}
	\label{fig:mission_execution}
\end{figure}

Fig. \ref{fig:mission_execution} shows four representative snapshots from a real mission in progress. 

\section{Conclusions}
\label{s:conclusions} 


We have introduced an efficient and simple algorithm for bi-criteria path planning.   Our method has been extensively tested both on synthetic data and in the field, as a component of a real robotic system.  Unlike prior methods based on scalarization, our approach recovers the entire Pareto Front regardless of its convexity.  As an additional bonus, choosing a larger value of $\delta$ boosts the efficiency of the method, resulting in an approximation of PF, whose accuracy can be further assessed in real time.  

The computational cost of our methods is $O(n m)$ corresponding to the number of nodes in the budget-augmented graph.  
Of course, the total number $p$ of nodes with distinct Pareto optimal cost tuples can be much smaller than $n m$.  
In such cases, the prior label-setting algorithms for multi-objective planning will likely be advantageous since their asymptotic cost is typically $O(p \log p)$.  
However, in this paper we are primarily interested in graphs used to approximate the path planning in continuous domains with edge costs correlated with the geometric distances. 
As we showed in Section \ref{s:bench}, for such graphs the number of points on PF is typically quite large (particularly as PRM graphs are further refined), with new Pareto optimal paths becoming feasible on most of the budget levels.  
Coupled with domain restriction techniques and the simplicity of implementation, this makes our approach much more attractive.

Several extensions would  obviously greatly expand the applicability of our method.  We hope to extend it to a higher number of simultaneous criteria and introduce A*, D*, and ``anytime planning'' versions.  In addition, it would be very useful to develop a priori bounds for the errors introduced in the PF as a function of $\delta$.  Another direction is automating the choice of primary/secondary cost to improve the method's efficiency. 

\textbf{Acknowledgements.} 
The authors would like to acknowledge and thank the whole Autonomy team at United Technologies Research Center.  
In particular, the authors would like to thank Suresh Kannan, Alberto Speranzon, and Andrzej Banaszuk for productive discussions of the algorithm and implementations of the simulation software.  
In addition, the authors would like to thank Amit Surana, Zohaib Mian, and Jerry Ding for the support from the Autonomy Initiative and help during field experiments. 
The authors are also grateful to Blane Rhoads for his suggestions on testing the slackness of the Pareto Front approximation.

\bibliographystyle{amsplain}
\bibliography{multio_PRM}

\end{document}